\documentclass[10pt,twocolumn,letterpaper]{article}

\usepackage{cvpr}              
\definecolor{cvprblue}{rgb}{0.21,0.49,0.74}
\usepackage[pagebackref,breaklinks,colorlinks,allcolors=cvprblue]{hyperref}
\usepackage{bm}
\usepackage{fontawesome}
\usepackage{nicefrac}
\def\paperID{1248} 
\def\confName{CVPR}
\def\confYear{2026}
\usepackage{pifont}
\usepackage{colortbl}
\usepackage{multirow}
\usepackage{multicol}
\usepackage{bbding} 

\title{FedMPT: Federated Multi-Label Prompt Tuning of Vision-Language Models}

\author{
Xucong Wang$^{1}$, Pengkun Wang$^{1,2,*}$, Zhe Zhao$^{1,3}$, Liheng Yu$^{1}$, Shuang Wang$^{1}$, Yang Wang$^{1,2,}$\thanks{Corresponding Author.} \\
$^1$ University of Science and Technology of China (USTC)\\
$^2$ Suzhou Institute for Advanced Research, USTC $^3$ City University of Hong Kong\\ 
{\tt\small \{xuco,zz4543,yuliheng,ws20021002\}@mail.ustc.edu.cn, \{pengkun,angyan\}@ustc.edu.cn } 
}

\begin{document}
\maketitle 

\begin{abstract}
Multi-Label Recognition (MLR) based on Vision-Language Models (VLMs) aims to leverage their pre-trained knowledge  to better adapt complex recognition scenarios, thereby enhancing model robustness. However, for realistic decentralized applications  requiring federated learning, adapting VLMs to each client that possesses private and heterogeneous data can cause the model to overfit spurious label correlations, consequently triggering irrelevant categories when encountering new samples. To tackle this problem, we reconsider the federated learning for MLR with a causal model, in which we adopt a front-door adjustment and decouple the MLR modeling process by intermediate variables that magnify the oracle label co-occurrence. Guided by our analysis, we propose our FedMPT, the first method specifically designed for federated MLR. The core idea of FedMPT is to leverage generalizable conditions to steer federated MLR to mitigate erroneous label activations. To achieve this, FedMPT introduces an Large Language Model (LLM)-driven pipeline to decipher the underlying conditions that govern label dependencies. Furthermore, we introduce an optimal transport between the condition-enriched prompts and the image patches to uncover multiple region-level semantics. Finally, we generate synergistic predictions from different conditions with a crafted gating mechanism. Experiments on multiple benchmark datasets show that our proposed approach achieves competitive results and outperforms SOTA methods under varied settings. Project Page: \href{https://xuc865.github.io/fedmpt/index.html}{https://xuc865.github.io/fedmpt/index.html}.
\end{abstract}

\begin{figure}[t] 
    \centering
    \includegraphics[width=1\linewidth]{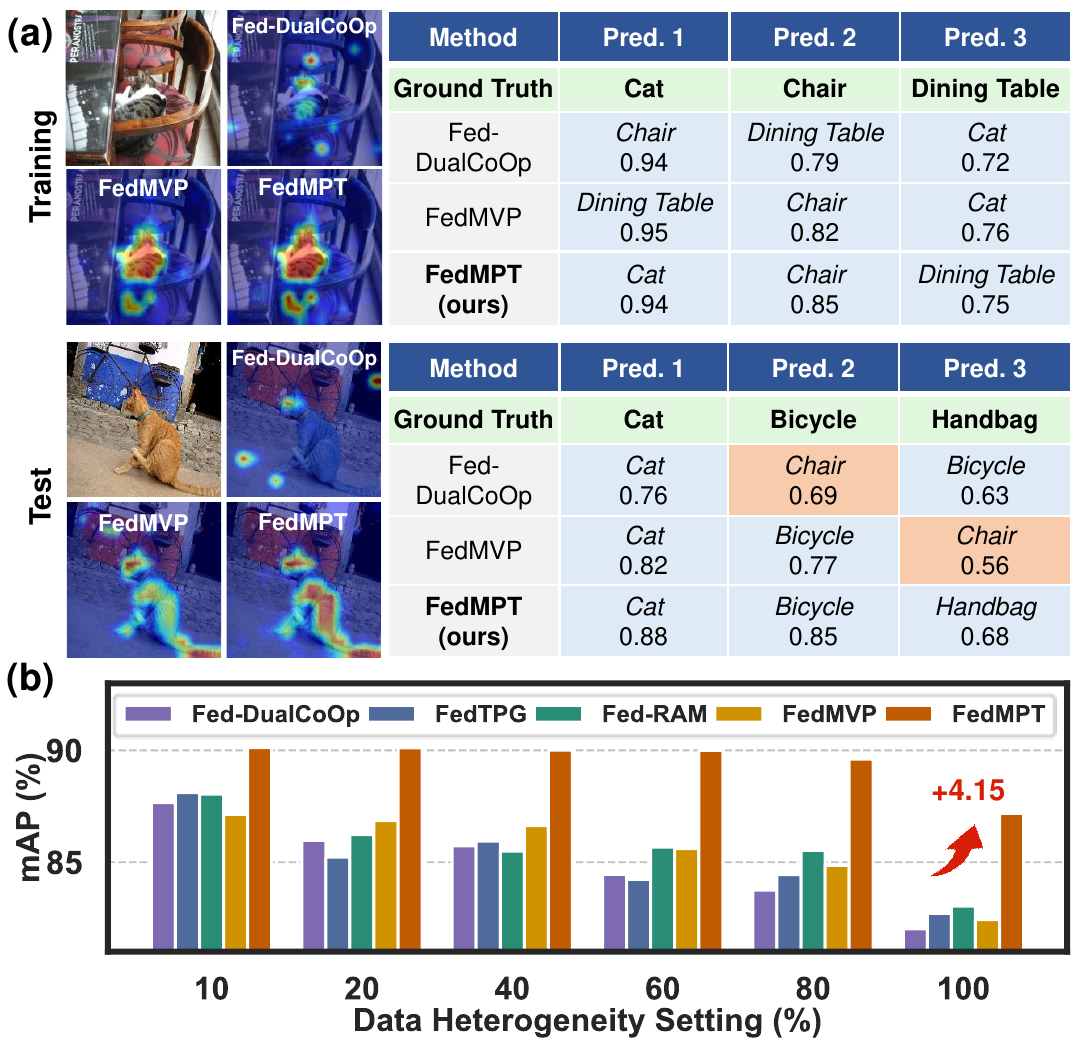} 
    \vspace{-0.6cm}
    \caption{\textbf{(a):} Comparison of class-activation map for ``\texttt{Cat}'' and top-3 predictions on the training image \textbf{(a, upper)} and test image  \textbf{(a, lower)}. Existing SOTAs are prone to overfitting spurious correlation (i.e., \texttt{cat-chair}) and diverting attentions under FL, while our FedMPT effectively alleviates these issues. \textbf{(b):} As data heterogeneity increases, existing SOTAs show significant degradation, while our FedMPT demonstrates substantial robustness. } 
    \vspace{-0.5cm}
    \label{fig:0}
\end{figure}  

\section{Introduction}
\label{sec:intro}
Multi-Label Recognition (MLR) aims to identify all possible labels in a single image.  Owing to its alignment with real-world requirements, MLR has found wide application~\cite{zhang2024multi,chang2020weakly,chen2020label,huang2024classification}. Early methods primarily focused on modeling inter-label co-occurrences~\cite{chen2019multi1,chen2019multi,durand2019learning,wang2020multi}, refining the models' attention on local regions~\cite{you2020cross,gao2021learning} or balancing the positive-negative gradients~\cite{ridnik2021asymmetric}. Recently, emerging efforts incorporate prompting pretrained Vision-Language Models (VLMs)~\cite{radford2021learning,liu2023visual,jia2021scaling,li2022blip,li2023blip} for MLR  owing to their remarkable zero-shot generalization abilities learned from web-scale image-text pairs. For instance, DualCoOp~\cite{sun2022dualcoop} and PosCoOp~\cite{rawlekar2025positivecoop} adapt the prompt learning to MLR by learning  two collaborative prompts for each class; ML-VPT~\cite{ma2025correlative} introduces distinct prompts for correlative and distinctive classes; SPARC~\cite{miller2025sparc} and CCD~\cite{kim2025classifier} unravel the inherent bias of VLMs to MLR as uneven score distribution across labels, then mitigate this bias for enhanced zero-shot learning and knowledge distillation respectively.

Although most MLR methods are designed for centralized settings where the model has  full access to the dataset space, real-world applications often necessitate a decentralized architecture~\cite{mcmahan2017communication,smith2017federated,zhao2018federated,cui2024harmonizing,singha2025fedmvp} (i.e., Federated Learning, FL) where each client only possesses a heterogeneous and private portion of the data. The long-standing research  focuses of FL (with VLMs) lies on modeling the commonality and specificity of client  distributions, for example, FedCoOp~\cite{guo2023promptfl}, PromptFL~\cite{guo2023promptfl} and FedTPG~\cite{qiu2024federated} introduce shared prompts with FedAvg~\cite{mcmahan2017communication} to learn generic knowledge across clients; FedPGP~\cite{cui2024harmonizing} and FedOTP~\cite{li2024global} introduces the local-global prompt collaboration for balancing generic and customized modeling. \textit{\textbf{Notably, to the best of our knowledge, all existing VLM-based FL methods are built for single-label recognition and consistently overlook the practical challenges of MLR}}.

Diving into the integration of MLR and FL, we present a critical two-fold \textit{\textbf{dilemma}}: first, if we directly train MLR SOTAs~\cite{rawlekar2025positivecoop,tan2025recover,singha2025fedmvp} on each client and aggregate their weights via FedAvg, the global model would learn excessively spurious~\cite{ma2025correlative} label correlations and show severe performance  degradation under increasing data heterogeneity~\cite{shi2025fedawa}. Second, conventional FL methods are ill-suited to MLR, as they fail to capture the inherent inter-dependencies between labels, similarly resulting in correlation overfitting and incomplete retrieval. Figure \ref{fig:0}.a illustrates  an example: the aggregated global model of existing SOTAs (DualCoOp~\cite{sun2022dualcoop} and FedMVP~\cite{singha2025fedmvp}) overfits to the \texttt{cat-chair} correlation, spuriously boosting the \texttt{chair} score upon seeing a \texttt{cat} in inference, meanwhile diverting their prediction confidence to ground-truth labels. Figure \ref{fig:0}.b summarizes mAP of different methods under varied data heterogeneity across clients (induced via clustering). We observe that existing SOTAs unanimously show sharp degradation despite their strong performance in near-IID settings (i.e., 10\%). 

To further understand MLR under FL, we reconsider it with a Structural Causal Modeling (SCM) in Section \ref{sec32}. Our key findings are that semantic variables learned from pre-training and control the content of images and labels, can be naturally divided into generic and client-specific variants. Overfitting the latter would lead to degraded generalization. Then, from the perspective of front-door adjustment, our objective is to identify an intermediate variable that maximizes the oracle label correlations.

Guided by our analysis, we propose our FedMPT, a novel condition-driven framework specifically designed for MLR under FL. FedMPT is built on a foundational idea: to leverage multiple, complementary conditions to intervene the MLR tasks. Concretely, we first devise an LLM-driven pipeline to generate generic abstract condition templates, which are incorporated into prompts for soft prompt learning; These condition prompts are then aligned with relevant image regions via optimal transport to produce multiple diverse, condition-specific predictions. Finally, inspired by the expert routing mechanism in LLMs~\cite{chen2023adamv,cai2025survey}, we introduce an adaptive gating mechanism to automatically adjust condition contributions in each client. We incorporate the Asymmetric Loss (ASL) as our training objective. In one communication round, all clients share their learnable parameters with the server, which aggregates them via FedAvg to form a unified global model. Comprehensive evaluations on three MLR benchmarks across various federated settings demonstrate that FedMPT substantially outperforms existing SOTA methods and exhibits remarkable robustness. Additional analyses validate the efficiency of our approach. Our contribution can be summarized as:
\begin{itemize}
    \item We identify and formalize the novel problem of Multi-Label Recognition (MLR) under realistic federated scenarios and unveil the vulnerability of existing methods.
    \item From a causal perspective, we attribute the intricacy of MLR under FL as the overfitting to local distributions and label correlations. Guided by our analysis, we propose FedMPT, which leverages multiple conditions to synergistically learn generic semantics across clients.
    \item Extensive experiments on three benchmarks show that FedMPT achieves state-of-the-art performance and exhibits remarkable robustness under various federated settings. Further ablations to highlight its efficiency.  
\end{itemize}

\section{Related Work}
\label{sec:rel} 

\textbf{Multi-Label Recognition (MLR).}
Multi-Label Recognition (MLR) demands the precise identification of all relevant labels in an image and naturally has broad real-world applications. Traditional MLR methods approaches have primarily focused on modeling class specifications and correlations; One line of work~\cite{chen2019multi,chen2019learning,wang2016cnn} incorporate text embedding graphs of labels to model the similarity of classes; Another line~\cite{ren2017multiple,huynh2020shared,narayan2021discriminative,ma2024text} dives into the local regions and discover the class-specific visual cues from each crop; More recently, the advancement of Vision-Language Models (VLMs) like CLIP has inspired a new direction for MLR: For instance, CDUL~\cite{abdelfattah2023cdul} devises an unsupervised framework where global and local knowledge are fused to generate pseudo labels for unlabeled samples; Concurrently, SPARC~\cite{miller2025sparc} and CCD~\cite{kim2025classifier} identify the inherent class bias in VLMs and explicitly mitigates this bias to achieve superior performance. Another notable progress lies in incorporating prompt learning to VLM-MLRs, for example, DualCoOp~\cite{sun2022dualcoop} and DualCoOp++~\cite{hu2023dualcoop++} introduce two prompts to model the object existence/non-existence in each image patch; RAM~\cite{tan2025recover} introduces a local knowledge guided aggregation scheme for open-vocabulary MLR; PosCoOp~\cite{rawlekar2025positivecoop} enhances DualCoOp with an unconditioned  prompt for object absence modeling.

\noindent\textbf{Federated Learning (FL) with VLMs.}
Federated Learning (FL)~\citep{mcmahan2017communication,smith2017federated,zhao2018federated,singha2025fedmvp} has emerged as a pivotal paradigm for enabling decentralized and privacy-preserving training on heterogeneous data. The application of FL to VLMs has seen significant evolution. Initial methods introduce prompt learning on each client with FedAvg to aggregate prompt weights~\citep{guo2023promptfl}; Subsequent research incorporates granular learnable modules like adapters~\cite{lu2023fedclip} or prompt generators~\cite{qiu2024federated,deng2024unlocking,singha2025fedmvp}. For example, FedMVP~\cite{singha2025fedmvp} generates visual embeddings with image tokens and LLM-attribute embeddings through a specialized cross-modal PromptFormer; Some other methods endeavor to harmonize the local and global knowledge~\cite{li2024global,cui2024harmonizing}, for example, FedOTP~\cite{li2024global} restrains the contribution of local/global prompts via margin-adapted optimal transport. Concurrently, FL is being explored in various specialized domains, including continual learning~\citep{yu2024personalized,zhang2025pfedmxf}, test-time adaptation~\cite{jiang2022test,bao2025latte}, autonomous driving~\citep{kou2025pfedlvm}, and interpretability~\citep{liuunderstanding2025}.  

\section{Preliminaries and Problem Analysis}
\subsection{Preliminaries}\label{sec31}

\paragraph{Multi-Label Recognition (MLR) with VLMs.} We first summarize  a baseline~\cite{sun2022dualcoop,rawlekar2025positivecoop} for MLR with VLMs built on prompt learning: given a typical VLM (CLIP) that employs dual encoders for processing multi-modal information, let $\mathcal{E}_{v}$ and $\mathcal{E}_{t}$ denote the image and text encoder respectively; Given the input $(\bm{x},\bm{y})$ where label  $\bm{y}\in\mathbb{R}^{C}$ ($C$ is the number of classes), CLIP encodes $\bm{x}$ into $M$-length visual embeddings $\bm{v}^0$ with $\mathcal{E}_{v}$; for the text modality, we fill all classnames into learnable templates (initialized as \textit{A photo of a [CLASS]}), yielding prompt $\bm{p}$; $\bm{p}$ is encoded into text embeddings $\bm{t}$ with $\mathcal{E}_{t}$. The operations at layer $i$ of  $\mathcal{E}_{v}$, $\mathcal{E}_{t}$ are:
{\small\begin{equation}
\begin{split}
   [\texttt{cls}^{i}, \bm{v}^{i}] = \mathcal{E}_{v}^{i}([\texttt{cls}^{i-1},\bm{v}^{i-1}]), \bm{v}^{0}=Emb(\bm{x}) \\ [\texttt{bot}^{i},\bm{p}^{i},\texttt{eot}^{i}] = \mathcal{E}_{t}^{i}([\texttt{bot}^{i-1},\bm{p}^{i-1},\texttt{eot}^{i-1}]),
\end{split}
\end{equation}}
where $[\cdot,\cdot]$ means concatenation, $\texttt{cls},\texttt{bot},\texttt{eot}$ represent the cls (class), bot (begin-of-text) and eot (end-of-text) tokens. $Emb$ is the patch embedding layer, $\bm{v}^{i}$ is the patch embeddings of layer $i$. The output projection $P_{t}$ is applied to $\texttt{eot}^{L}$ to generate the text embedding, i.e., $\bm{f}_{t}=P_{t}(\texttt{eot}^{L})$. For the visual modality, instead of using the global representation $\texttt{cls}^{L}$, this baseline projects $\bm{v}^{L}$ into the final patch-level output embeddings $\bm{f}_{v}(\bm{v})$, i.e., $\bm{f}_{v}(\bm{v})=P_{v}(\bm{v}^{L})$. The final prediction is calculated by selectively aggregating the predictions over patches (similarity between $\bm{f}_{v}(\bm{v})$ and $\bm{f}_{t}$) based on their softmax-normalized weights:
{\small\begin{equation}
    \mathbb{P}(y_{c}|\bm{x})=\sum_m\frac{\exp({s}_{m,c}/\tau)}{\sum_{c'}\exp({s}_{m,c'}/\tau)}\cdot{s}_{m,c},
\end{equation}}
where $\texttt{sim}(\cdot,\cdot)$ represents the cosine similarity in default, ${s}_{m,c}=\texttt{sim}(\bm{f}_{v}(\bm{v}_m),\bm{f}_{t,c})$ is the cosine similarity between patch $m$ and class $c$. $\tau$ is the temperature.  The final logits are optimized with the Asymmetric Loss (ASL)~\cite{ridnik2021asymmetric} to handle optimization imbalance of positive and negative classes:
{\small\begin{equation}
    \mathcal{L}_{asl}=(1-\mathbb{P})^{\gamma_+}\bm{y}\log(\mathbb{P})+(\mathbb{P}^c)^{\gamma_-}(1-\bm{y})\log(1-\mathbb{P}^c)
\end{equation}}
where $\mathbb{P}^c = \max(\mathbb{P}-c,0)$ is for truncating negative predictions, which is controlled by the hard threshold $c$. We set the hyper-parameters as $\gamma_{-}\ge\gamma_{+}$, so that ASL would better down-weight the contribution of easy negative samples.
\vspace{-0.1cm}
\paragraph{Federated Learning (FL).} Following previous approaches~\cite{qiu2024federated,singha2025fedmvp}, we consider a standard FL system comprising $K$ remote client models $\{\bm{\rho}_{k}\}_{k=1}^{K}$ running optimization on their local data $D_{k}$, as well as a server $\mathcal{G}$ for coordination by aggregating and broadcasting parameters. We follow a non-IID federated setup where local data for different client are heterogeneous; \textit{to achieve this under MLR settings, we cluster the dataset based on the image features extracted from zero-shot ViT/B-16, then assign each cluster to one client.} The objective is to learn an optimal global model $\bm{\rho}$ aggregated from clients with the minimum risk $\mathcal{L}$, undergoing $\phi$ communication rounds with a client participation rate of $\epsilon$:
{\small\begin{equation}
\mathcal{L}=\texttt{min}_{\bm{\rho}}\sum\nolimits_{k=1}^{K}p_{k}\mathcal{L}_{k}(\bm{\rho},\mathcal{D}_{k};\mathcal{G}[{\phi};\epsilon]),
\end{equation}}
where $p_{k}$ represents the weight of $k$-th client and set to $\nicefrac{|\mathcal{D}_{k}|}{\sum_{\mathcal{D}_{l}\in\mathcal{D}_{k}}|\mathcal{D}_{l}|}$, where $|\mathcal{D}_{k}|$ is the size of $\mathcal{D}_{k}$.

\subsection{Problem Analysis}\label{sec32}
\begin{figure}[t] 
    \centering
    \includegraphics[width=0.95\linewidth]{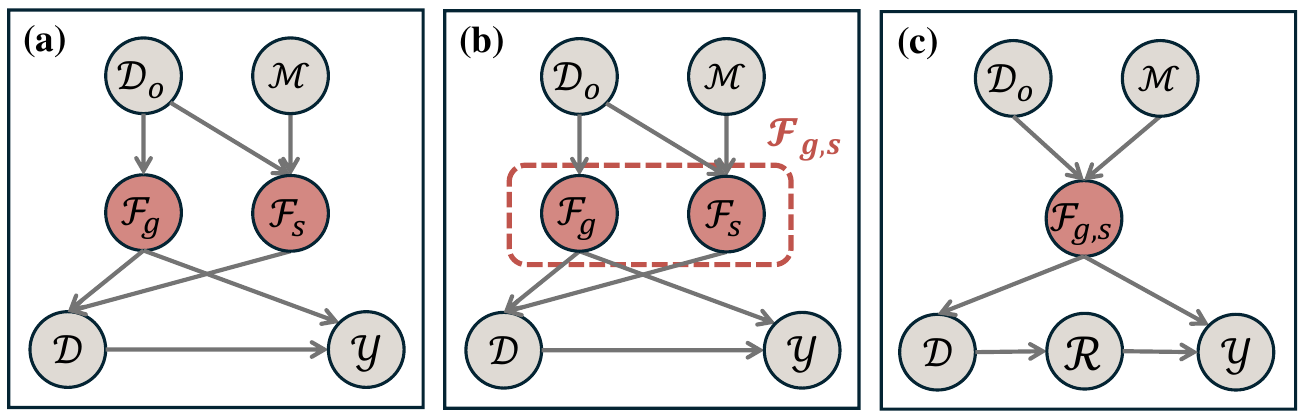} 
    \vspace{-0.2cm}
    \caption{Structural Causal Model (SCM) for MLR under FL.}  
    \label{fig:1}
     \vspace{-0.3cm}
\end{figure}

This subsection formalizes the problem of MLR under FL from a causal perspective. Our proposed Structural Causal Model (SCM) is depicted in Figure \ref{fig:1}, where nodes represent variables during pre-training or fine-tuning, and edges denote causal relationships. Unobservable and observable variables are highlighted in red and gray, respectively.

Concretely, $\mathcal{D}_o$ means the pre-training data, which determines the semantic factors $\mathcal{F}$ which controls the semantics of input space $\mathcal{D}$ and output space $\mathcal{Y}$. As shown in Figure \ref{fig:1}.a, under the federated learning scenario, $\mathcal{F}$ can be divided into generic factors $\mathcal{F}_{g}$ which capture transferable knowledge across clients and are the target of finetuning, and $\mathcal{F}_{s}$, which encapsulate client-specific semantics that may induce overfitting to local spurious correlations. Notably, $\mathcal{F}_{s}$ is also influenced by manual factors $\mathcal{M}$ like data partitioning policies. The image content is a mixture of both, whereas the labels can only be derived from  the generic factors $\mathcal{F}_{g}$.

Our objective is to maximize the influence of $\mathcal{F}_{g}$ while minimizing that of $\mathcal{F}_{s}$ during training. This enables an unbiased estimation of the causal effect $\mathcal{D} \rightarrow \mathcal{Y}$. However, the insufficiency of local training data and its dramatic gap with inference data makes our modeling biased and capturing a collection $\mathcal{F}_{g,s}$ of both $\mathcal{F}_{g}$ and $\mathcal{F}_{s}$ (Figure \ref{fig:1}.b), resulting in a backdoor-path of $\mathcal{D}\leftarrow\mathcal{F}_{g,s}\rightarrow\mathcal{Y}$. To tackle  this issue, we incorporate a well-known front-door adjustment~\cite{wen2025causal,zhang2024rethinking} by introducing an intermediate variable $\mathcal{R}$ (Figure \ref{fig:1}.c) that ideally reflects the causality between $\mathcal{D}$ and $\mathcal{Y}$. We can then identify $P(\mathcal{Y}|do(\mathcal{D}))$ with front-door adjustment:
{\small\begin{equation}
P(\mathcal{Y}|do(\mathcal{D}))=\mathbb{E}_{P(r|d)}\mathbb{E}_{P(d')}P(Y|r,d')
\end{equation}}
where $d,r$ denote specific values of $\mathcal{D}$ and $\mathcal{R}$. The core challenge thus reduces to constructing $r$ that accurately captures the oracle causal mechanism $\mathcal{D}\rightarrow\mathcal{Y}$. In the next section, we'll introduce our FedMPT, which incorporates condition-guided learning and gating to meet our analysis.

\section{Methodology}
This section introduces our proposed FedMPT, comprising condition prompt generation (\S 4.1), condition-guided optimal transport (\S 4.2), condition gating (\S 4.3), and the federated communication process (\S 4.4).

\subsection{Condition Prompt Generation}
\textit{How can we maximize the adjustment of variable $r$ of SCM in \S 3.2?} Since directly learning from the datasets leads to spurious correlations and degraded generalization, we propose to intervene MLR with certain \textbf{\textit{conditions}} that approximate the oracle causalities and label correlations: Recalling the instance of Figure \ref{fig:0}.a, we tend to accept the cat-chair concurrent predictions under conditions of \textit{``indoor scene''}, \textit{``wooden textures''}, and \textit{``lying actions''}; Thus, the model will reduce \texttt{chair}'s weight when faced with a (cat, bicycle) image, where some \textbf{\textit{conditions}} are not satisfied.

\begin{figure}[t] 
    \centering
    \includegraphics[width=1\linewidth]{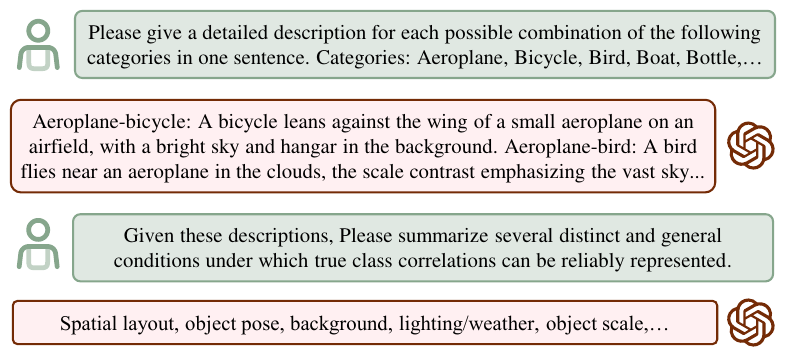} 
    \vspace{-0.6cm}
    \caption{Our proposed LLM-based condition generation pipeline.}  
    \label{fig:2}
     \vspace{-0.5cm}
\end{figure}

Pursuing this idea, our goal is to generate a set of generic, broad, and fine-grained conditions that can be shared across all clients. Our strategy is to fix some abstract conditions and leave the specific and contextualized contents learnable. To generate the abstract conditions, we employ an LLM-driven pipeline (Figure \ref{fig:2}) with Chain-of-Thought (CoT) following~\cite{li2024advancing}. Concretely, we first prompt the LLM to generate as many descriptions as possible for each possible combination of dataset categories; thanks to the rich knowledge embedded in LLMs, we can acquire the characteristics and existence conditions of various label combinations at this stage. Then, we prompt LLM to summarize $N$ distinct abstract conditions which would exactly encapsulate label correlations; finally, we obtain several abstract conditions like \textit{``spatial layout''}, \textit{``object pose''}, \textit{``background''}, etc. 

To integrate these abstract conditions into the learning process, we populate them into the  ``[COND]'' slot of the following template, yielding prompts $\bm{p}^{\dagger}=\{\bm{p}^{\dagger}_1,...,\bm{p}^{\dagger}_C\}$:
\begin{center}
    [$L^{}_1$]$\cdots$[$L_{\beta_{cond}}$] [COND]  [$L_1$]$\cdots$[$L_{\beta_{cls}}$] [CLASS],
\end{center}
where $[L_{\cdot}]$ means the learnable tokens, $\beta_{cond}$, $\beta_{cls}$ control the number of condition-level and class-level tokens respectively. Critically, the former ones are specified for each condition, while the latter ones are shared by all classes. $\bm{p}^{\dagger}$ is maintained in the server and distributed to all clients in the communication phase. We denote $\bm{f}_t(\bm{p}^{\dagger})$ as the output text embeddings of $\bm{p}^{\dagger}$ processed by the text encoder. Concrete conditions and more discussions are in \textbf{\underline{Sup. Mat. D}}.
 
\begin{figure*}[t] 
    \centering
    \includegraphics[width=1\linewidth]{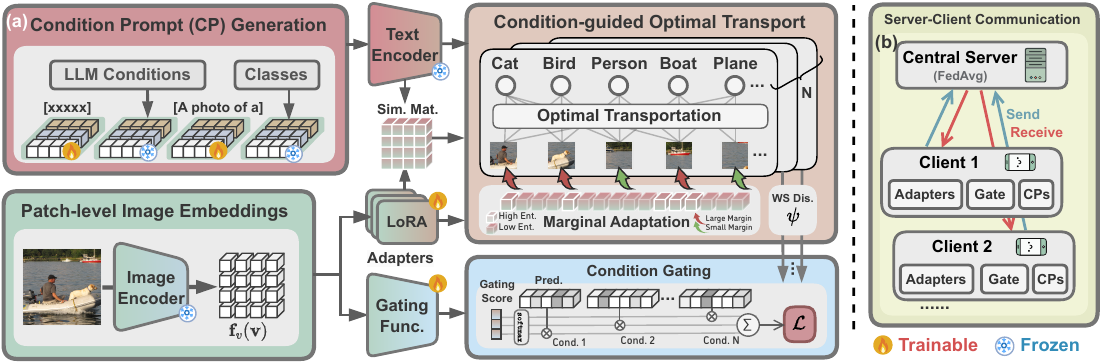} 
    \vspace{-0.6cm}
    \caption{Overview of our proposed FedMPT framework. (a) The LLM-generated conditions are instantiated into Condition Prompts (CPs), which are encoded into text embeddings. For a given image, its visual feature map is aligned with these prompt embeddings via Optimal Transport (OT). The contributions of different conditions are then adaptively calibrated by a gating module. (b) At each communication 
    round, the server aggregates the parameters of CPs, adapters, and gating modules and distributes the updated parameters back. } 
    \vspace{-0.4cm}
    \label{fig:main}
\end{figure*}  

\subsection{Condition-guided Optimal Transport}
To better align the generated condition prompts with region-level fine-grained visual semantics, we devise an Optimal-Transport (OT) between output patch embeddings (denoted as $\bm{f}_{v}(\bm{v})$ in the above) and the condition prompts received from the server. Specifically, we first introduce $N$  adapters $\{\mathcal{A}_n\}_{n=1}^{N}$ (corresponding to attributes) over $\bm{f}_{v}(\bm{v})$ to generate new visual latent spaces, where each adapter is a LoRA-like~\cite{tian2024hydralora,yang2024mma} architecture for efficiency:
{\small\begin{equation}
   \bm{f}^{\dagger}_{v,n}(\bm{v}) = \mathcal{A}_n(\bm{f}_{v}(\bm{v}))=W_{\uparrow}(W_{\downarrow}(\bm{f}_{v}(\bm{v}))),
\end{equation}}
where $W_{\downarrow}\in\mathbb{R}^{D\times D_s}$,$W_{\uparrow}\in\mathbb{R}^{D_s\times D}$, $D$/$D_s$ is the output / down-projected latent dimension.
The OT aims to find an optimal plan $\mathcal{P}^{*}\in\mathbb{R}^{M\times N\times C}$ that minimizes the distance between distributions, i.e., $\mathcal{P}^{*}=\texttt{OT}(\mathcal{C};\bm{a},\bm{b})$, where $\mathcal{C}\in\mathbb{R}^{M\times N\times C}$ represents the cost-matrix, $\bm{a}\in\mathbb{R}^{N}$, $\bm{b}\in\mathbb{R}^{M}$ are constrained marginal distributions.  $\mathcal{C}$ is calculated with:
{\small\begin{equation}
        \mathcal{C}_{m,n}=1-\frac{\exp(\texttt{sim}(\bm{f}^{\dagger}_{v,n}(\bm{v}_m),\bm{f}_t(\bm{p}_n^{\dagger}))/\tau)}{\sum_m\exp(\texttt{sim}(\bm{f}^{\dagger}_{v,n}(\bm{v}_m),\bm{f}_t(\bm{p}_n^{\dagger}))/\tau)}.  
\end{equation}}
We also denote $\mathcal{S}=1-\mathcal{C}$, i.e., the original region-text similarity. We keep $\bm{b}$ as the uniform distribution so that all categories yield equal change to be detected in the image. For $\bm{a}$, inspired by recent finding~\cite{chen2025interpretable} indicating that regions yield different contributions of semantics, we set $\bm{a}$ as the semantic importance of each patch, calculated by:
\begin{equation}\small
    {a}_{m,n}=\frac{\exp(\max(\texttt{sim}(\bm{f}^{\dagger}_{v,n}(\bm{v}_m),\bm{f}_t(\bm{p}^{\dagger}_n)))/\tau)}{\sum_m \exp(\max(\texttt{sim}(\bm{f}^{\dagger}_{v,n}(\bm{v}_m),\bm{f}_t(\bm{p}^{\dagger}_n)))/\tau)},
\end{equation}
where $\max$ is applied to the class-dimension of the calculated similarities. To calculate OT more efficiently, we introduce entropy relaxation to approximate the results with the Sinkhorn~\cite{sinkhorn1967concerning} algorithm, formulated as:
\begin{equation}\small 
    \mathcal{P}\!=\!\texttt{diag}(\mathcal{U})\mathcal{M}\texttt{diag}(\mathcal{V}),  \ \mathcal{U}=\{u_{m}\}_{m=1}^{M}, \mathcal{V}\text{=}\{v_{n}\}_{n=1}^{N},
\end{equation}
where $\mathcal{U}$, $\mathcal{V}$ are updated with the following recurrent form:
\begin{equation}\small
    u_{m}\!\!\leftarrow\!\!\frac{\bm{a}}{\sum_n\mathcal{M}_{m,n}v_n}, v_{n}\!\!\leftarrow\!\!\frac{\bm{b}}{\sum_n\mathcal{M}_{m,n}u_n}, \mathcal{M}\text{=}\exp(\frac{-\mathcal{C}}{\lambda})
\end{equation}
With the OT applied across all classes for each conditioned prompt, we finally obtain $N$ similarity maps between regions and classes. For each client, we calculate the Wasserstein distance $\bm{\psi}\in\mathbb{R}^{N\times C}$ by class for each condition, which reflects the affinity of each class to the visual regions and calculated via $\bm{\psi}_{n} = \sum_{m}\mathcal{P}_{m,n}\mathcal{S}_{m,n}$. Consequently, we treat $\bm{\psi}$ as conditioned predictions, i.e., $\mathbb{P}_{n}=\bm{\psi}_{n}$.

\subsection{Condition Gating} 
While the conditional prompting and OT matching mitigate overfitting to local spurious correlations in MLR, the relevance of each condition may not always remain the same across clients due to data heterogeneity. To ensure robust generalization, we introduce a gating mechanism that dynamically adapts the influence of each condition. Specifically, inspired by Mixture-of-Experts (MoE)~\cite{tian2024hydralora,wu2025routing} in LLMs, we leverage a router $\omega\in\mathbb{R}^{D\times N}$ to dynamically determine the contribution of  different conditions and aggregate their predictions:
\begin{equation}\small
    \omega=\Omega(\bm{f}_v(\bm{v})); \ \ \ \mathbb{P}'=\sum\nolimits_{n}\frac{\exp(\omega_n)}{\sum_{n'}\exp(\omega_{n'})}\mathbb{P}_{n},
\end{equation}
where $\Omega\in\mathbb{R}$ is a similar LoRA module like $\{\mathcal{A}_i\}$.

\subsection{Local Training and Federated Average} 
\textbf{Local Training.} Each client optimizes the local model using their private data   via the above asymmetric loss (ASL):
\begin{equation}\small
        \mathcal{L}=(1-\mathbb{P}')^{\gamma_+}\bm{y}\log(\mathbb{P}')+(\mathbb{P}'_c)^{\gamma_-}(1-\bm{y})\log(1-\mathbb{P}'_c)
\end{equation}
Recall that $\gamma_+$,$\gamma_-$ are hyper-parameters to control the contribution of positive/negative regularizations. Ablations of these hyper-parameters are in \textbf{\underline{Sup. Mat. B.}}

\noindent\textbf{Federated Average.} In one communication round, the server collects the updated weights of condition prompts $\bm{p}$, adapters $\{\mathcal{A}_n\}_{n=1}^{N}$ and gates $\Omega$ from clients, where they're aggregated among different clients $\mathtt{Cli}_t$ with FedAvg~\cite{mcmahan2017communication}:
{\small\begin{equation}
    \{\bm{p},\{\mathcal{A}_n\}_{n=1}^{N},\Omega\}\leftarrow\frac{1}{K}\sum\nolimits_{t}\mathtt{Cli}_{t}(\{\bm{p},\{\mathcal{A}_n\}_{n=1}^{N},\Omega\})
\end{equation}}
The aggregated weights are sent to clients for the subsequent training. The above steps are repeated for $R$ rounds.

\begin{table*}[t]
\caption{Results on the Heterogeneous Benchmark. We report the mAP, per-category F1 (CF1) and overall F1 (OF1) with the client number varies from 10\% to 100\% of the class number. The best results are marked with \textbf{bold}.}
\vspace{-0.5cm}
\begin{center} 
\renewcommand{\arraystretch}{1}   
\setlength\tabcolsep{2.5pt} 
\resizebox{1\textwidth}{!}{
\begin{tabular}{ccccccccccccccccccccccc}
\hline
\multicolumn{23}{c}{VOC2007} \\ \hline
\multicolumn{1}{c|}{} & \multicolumn{1}{c|}{} & \multicolumn{3}{c|}{$t=10\%$} & \multicolumn{3}{c|}{$t=20\%$} & \multicolumn{3}{c|}{$t=40\%$} & \multicolumn{3}{c|}{$t=60\%$} & \multicolumn{3}{c|}{$t=80\%$} & \multicolumn{3}{c|}{$t=100\%$} & \multicolumn{3}{c}{Avg} \\ \cline{3-23} 
\multicolumn{1}{c|}{\multirow{-2}{*}{Methods}} & \multicolumn{1}{c|}{\multirow{-2}{*}{Venue}} & mAP & CF1 & \multicolumn{1}{c|}{OF1} & mAP & CF1 & \multicolumn{1}{c|}{OF1} & mAP & CF1 & \multicolumn{1}{c|}{OF1} & mAP & CF1 & \multicolumn{1}{c|}{OF1} & mAP & CF1 & \multicolumn{1}{c|}{OF1} & mAP & CF1 & \multicolumn{1}{c|}{OF1} & mAP & CF1 & OF1 \\ \hline
\multicolumn{1}{c|}{Fed-DualCoOp} & \multicolumn{1}{c|}{NeurIPS'22} & 87.67 & 81.32 & \multicolumn{1}{c|}{81.91} & 85.98 & 78.84 & \multicolumn{1}{c|}{76.16} & 85.73 & 78.33 & \multicolumn{1}{c|}{76.44} & 84.46 & 77.13 & \multicolumn{1}{c|}{75.25} & 83.76 & 75.44 & \multicolumn{1}{c|}{73.89} & 82.02 & 75.00 & \multicolumn{1}{c|}{74.73} & 84.94 & 77.68 & 76.40 \\
\multicolumn{1}{c|}{Fed-SCPNet} & \multicolumn{1}{c|}{CVPR'23} & 83.37 & 78.74 & \multicolumn{1}{c|}{77.86} & 81.15 & 74.68 & \multicolumn{1}{c|}{76.05} & 80.03 & 74.05 & \multicolumn{1}{c|}{74.09} & 79.89 & 66.04 & \multicolumn{1}{c|}{73.45} & 76.51 & 74.29 & \multicolumn{1}{c|}{71.13} & 76.48 & 74.06 & \multicolumn{1}{c|}{70.64} & 79.57 & 73.64 & 73.87 \\
\multicolumn{1}{c|}{Fed-MaPLe} & \multicolumn{1}{c|}{CVPR'23} & 84.22 & 78.00 & \multicolumn{1}{c|}{82.36} & 87.87 & 81.27 & \multicolumn{1}{c|}{80.70} & 84.36 & 76.35 & \multicolumn{1}{c|}{75.82} & 81.82 & 73.73 & \multicolumn{1}{c|}{73.55} & 80.71 & 70.65 & \multicolumn{1}{c|}{72.04} & 76.08 & 68.13 & \multicolumn{1}{c|}{70.89} & 82.51 & 74.69 & 75.89 \\
\multicolumn{1}{c|}{FedPGP} & \multicolumn{1}{c|}{ICML'24} & 85.47 & 77.97 & \multicolumn{1}{c|}{79.81} & 84.51 & 76.76 & \multicolumn{1}{c|}{77.66} & 84.47 & 75.99 & \multicolumn{1}{c|}{73.97} & 79.89 & 72.72 & \multicolumn{1}{c|}{67.48} & 76.60 & 68.71 & \multicolumn{1}{c|}{66.52} & 78.16 & 69.30 & \multicolumn{1}{c|}{68.35} & 81.52 & 73.58 & 72.30 \\
\multicolumn{1}{c|}{Fed-TCP} & \multicolumn{1}{c|}{CVPR'24} & 81.56 & 76.02 & \multicolumn{1}{c|}{79.23} & 81.15 & 72.41 & \multicolumn{1}{c|}{79.50} & 80.65 & 78.93 & \multicolumn{1}{c|}{\underline{80.44}} & 77.32 & 78.38 & \multicolumn{1}{c|}{78.89} & 76.50 & 76.64 & \multicolumn{1}{c|}{73.50} & 76.42 & 71.52 & \multicolumn{1}{c|}{76.48} & 78.93 & 75.65 & 78.01 \\
\multicolumn{1}{c|}{FedTPG} & \multicolumn{1}{c|}{ICLR'24} & 88.11 & 81.13 & \multicolumn{1}{c|}{78.05} & 85.23 & 81.26 & \multicolumn{1}{c|}{77.73} & 85.95 & 79.05 & \multicolumn{1}{c|}{78.54} & 84.23 & 77.37 & \multicolumn{1}{c|}{78.68} & 84.45 & 77.29 & \multicolumn{1}{c|}{76.13} & 82.72 & 76.67 & \multicolumn{1}{c|}{75.87} & 85.12 & 78.80 & 77.50 \\
\multicolumn{1}{c|}{Fed-PosCoOp} & \multicolumn{1}{c|}{WACV'25} & 87.42 & 80.67 & \multicolumn{1}{c|}{79.51} & 84.71 & 82.77 & \multicolumn{1}{c|}{77.86} & 82.53 & \underline{80.90} & \multicolumn{1}{c|}{76.97} & 81.65 & 80.31 & \multicolumn{1}{c|}{76.43} & 80.71 & 80.17 & \multicolumn{1}{c|}{75.79} & 79.90 & 78.26 & \multicolumn{1}{c|}{75.08} & 82.82 & \underline{80.51} & 76.94 \\
\multicolumn{1}{c|}{FedAWA} & \multicolumn{1}{c|}{CVPR'25} & 84.75 & 78.57 & \multicolumn{1}{c|}{80.41} & 83.17 & 76.06 & \multicolumn{1}{c|}{78.35} & 81.77 & 75.11 & \multicolumn{1}{c|}{79.99} & 79.48 & 72.26 & \multicolumn{1}{c|}{79.24} & 78.81 & 70.49 & \multicolumn{1}{c|}{\underline{77.68}} & 78.22 & 70.73 & \multicolumn{1}{c|}{70.15} & 81.03 & 73.87 & 77.64 \\
\multicolumn{1}{c|}{Fed-RAM} & \multicolumn{1}{c|}{CVPR'25} & \underline{88.05} & \underline{81.50} & \multicolumn{1}{c|}{81.98} & 86.24 & \underline{81.27} & \multicolumn{1}{c|}{\underline{80.70}} & 85.50 & 79.74 & \multicolumn{1}{c|}{80.01} & \underline{85.68} & 79.24 & \multicolumn{1}{c|}{78.90} & \underline{85.53} & \underline{79.56} & \multicolumn{1}{c|}{72.84} & \underline{83.04} & 77.00 & \multicolumn{1}{c|}{68.99} & \underline{85.67} & 79.72 & 77.24 \\
\multicolumn{1}{c|}{FedMVP} & \multicolumn{1}{c|}{ICCV'25} & 87.14 & 80.24 & \multicolumn{1}{c|}{\underline{82.30}} & \underline{86.86} & 79.32 & \multicolumn{1}{c|}{80.22} & \underline{86.64} & 79.00 & \multicolumn{1}{c|}{79.93} & 85.61 & \underline{79.37} & \multicolumn{1}{c|}{\underline{80.65}} & 84.87 & 79.23 & \multicolumn{1}{c|}{75.60} & 82.43 & \underline{79.56} & \multicolumn{1}{c|}{\underline{77.92}} & 85.59 & 79.45 & \underline{79.44} \\
\rowcolor{gray!10}\multicolumn{1}{c|}{FedMPT} & \multicolumn{1}{c|}{Ours} & \textbf{90.13} & \textbf{84.33} & \multicolumn{1}{c|}{\textbf{86.82}} & \textbf{90.12} & \textbf{84.91} & \multicolumn{1}{c|}{\textbf{83.23}} & \textbf{90.01} & \textbf{84.42} & \multicolumn{1}{c|}{\textbf{84.31}} & \textbf{90.00} & \textbf{83.88} & \multicolumn{1}{c|}{\textbf{84.56}} & \textbf{89.61} & \textbf{83.72} & \multicolumn{1}{c|}{\textbf{80.36}} & \textbf{87.19} & \textbf{81.31} & \multicolumn{1}{c|}{\textbf{82.43}} & \textbf{89.51} & \textbf{83.76} & \textbf{83.62} \\
\rowcolor{gray!10}\multicolumn{1}{c|}{\textbf{$\Delta$ Prev. best}} & \multicolumn{1}{c|}{\textbackslash{}} & \textcolor{red}{\textbf{+2.02}} & \textcolor{red}{\textbf{+2.83}} & \multicolumn{1}{c|}{\textcolor{red}{\textbf{+4.46}}} & \textcolor{red}{\textbf{+2.25}} & \textcolor{red}{\textbf{+2.14}} & \multicolumn{1}{c|}{\textcolor{red}{\textbf{+2.53}}} & \textcolor{red}{\textbf{+3.37}} & \textcolor{red}{\textbf{+3.52}} & \multicolumn{1}{c|}{\textcolor{red}{\textbf{+3.87}}} & \textcolor{red}{\textbf{+4.32}} & \textcolor{red}{\textbf{+3.57}} & \multicolumn{1}{c|}{\textcolor{red}{\textbf{+3.91}}} & \textcolor{red}{\textbf{+4.08}} & \textcolor{red}{\textbf{+3.55}} & \multicolumn{1}{c|}{\textcolor{red}{\textbf{+2.68}}} & \textcolor{red}{\textbf{+4.15}} & \textcolor{red}{\textbf{+1.75}} & \multicolumn{1}{c|}{\textcolor{red}{\textbf{+4.51}}} & \textcolor{red}{\textbf{+3.84}} & \textcolor{red}{\textbf{+3.25}} & \textcolor{red}{\textbf{+4.18}} \\ \hline
\multicolumn{23}{c}{COCO2014} \\ \hline
\multicolumn{1}{c|}{} & \multicolumn{1}{c|}{} & \multicolumn{3}{c|}{$t=10\%$} & \multicolumn{3}{c|}{$t=20\%$} & \multicolumn{3}{c|}{$t=40\%$} & \multicolumn{3}{c|}{$t=60\%$} & \multicolumn{3}{c|}{$t=80\%$} & \multicolumn{3}{c|}{$t=100\%$} & \multicolumn{3}{c}{Avg} \\ \cline{3-23} 
\multicolumn{1}{c|}{\multirow{-2}{*}{Methods}} & \multicolumn{1}{c|}{\multirow{-2}{*}{Venue}} & mAP & CF1 & \multicolumn{1}{c|}{OF1} & mAP & CF1 & \multicolumn{1}{c|}{OF1} & mAP & CF1 & \multicolumn{1}{c|}{OF1} & mAP & CF1 & \multicolumn{1}{c|}{OF1} & mAP & CF1 & \multicolumn{1}{c|}{OF1} & mAP & CF1 & \multicolumn{1}{c|}{OF1} & mAP & CF1 & OF1 \\ \hline
\multicolumn{1}{c|}{Fed-DualCoOp} & \multicolumn{1}{c|}{NeurIPS'22} & 61.88 & 58.34 & \multicolumn{1}{c|}{62.15} & 58.39 & 54.11 & \multicolumn{1}{c|}{59.82} & 55.95 & 51.99 & \multicolumn{1}{c|}{57.17} & 53.80 & 49.55 & \multicolumn{1}{c|}{58.33} & 52.28 & 48.34 & \multicolumn{1}{c|}{57.86} & 51.46 & 48.57 & \multicolumn{1}{c|}{56.95} & 55.63 & 51.82 & 58.71 \\
\multicolumn{1}{c|}{Fed-SCPNet} & \multicolumn{1}{c|}{CVPR'23} & 60.96 & 57.96 & \multicolumn{1}{c|}{59.70} & 57.68 & 52.91 & \multicolumn{1}{c|}{57.45} & 52.79 & 53.09 & \multicolumn{1}{c|}{53.50} & 50.10 & 50.73 & \multicolumn{1}{c|}{55.36} & 48.40 & 48.84 & \multicolumn{1}{c|}{45.61} & 46.67 & 42.25 & \multicolumn{1}{c|}{47.75} & 52.77 & 50.96 & 53.23 \\
\multicolumn{1}{c|}{Fed-MaPLe} & \multicolumn{1}{c|}{CVPR'23} & 64.83 & \underline{61.28} & \multicolumn{1}{c|}{64.45} & 62.98 & 58.40 & \multicolumn{1}{c|}{62.16} & 61.80 & 58.25 & \multicolumn{1}{c|}{57.43} & 60.81 & \underline{57.83} & \multicolumn{1}{c|}{55.89} & 55.83 & 51.76 & \multicolumn{1}{c|}{52.54} & 47.24 & 49.97 & \multicolumn{1}{c|}{50.13} & 58.92 & \underline{56.25} & 57.10 \\
\multicolumn{1}{c|}{FedPGP} & \multicolumn{1}{c|}{ICML'24} & 63.10 & 61.09 & \multicolumn{1}{c|}{65.36} & 59.86 & 58.35 & \multicolumn{1}{c|}{64.78} & 58.21 & 57.52 & \multicolumn{1}{c|}{62.16} & 58.18 & 55.79 & \multicolumn{1}{c|}{60.87} & 56.52 & 52.36 & \multicolumn{1}{c|}{\underline{60.85}} & 54.53 & 53.19 & \multicolumn{1}{c|}{53.54} & 58.40 & 56.38 & 61.26 \\
\multicolumn{1}{c|}{Fed-TCP} & \multicolumn{1}{c|}{CVPR'24} & 62.66 & 61.50 & \multicolumn{1}{c|}{67.16} & 62.79 & 57.46 & \multicolumn{1}{c|}{62.69} & 61.77 & 53.16 & \multicolumn{1}{c|}{60.13} & 60.07 & 52.46 & \multicolumn{1}{c|}{57.72} & 59.99 & 50.93 & \multicolumn{1}{c|}{55.78} & 57.22 & 47.75 & \multicolumn{1}{c|}{53.56} & 60.75 & 53.88 & 59.51 \\
\multicolumn{1}{c|}{FedTPG} & \multicolumn{1}{c|}{ICLR'24} & 63.91 & 59.39 & \multicolumn{1}{c|}{64.70} & 61.34 & 56.05 & \multicolumn{1}{c|}{55.72} & 60.95 & 55.47 & \multicolumn{1}{c|}{51.31} & 60.54 & 54.52 & \multicolumn{1}{c|}{54.18} & 54.53 & 51.63 & \multicolumn{1}{c|}{53.92} & 50.88 & 53.18 & \multicolumn{1}{c|}{52.97} & 58.69 & 55.04 & 55.47 \\
\multicolumn{1}{c|}{Fed-PosCoOp} & \multicolumn{1}{c|}{WACV'25} & 63.28 & 58.69 & \multicolumn{1}{c|}{62.57} & 59.94 & 55.71 & \multicolumn{1}{c|}{60.91} & 55.88 & 50.41 & \multicolumn{1}{c|}{60.13} & 53.23 & 49.58 & \multicolumn{1}{c|}{59.37} & 50.95 & 50.03 & \multicolumn{1}{c|}{58.04} & 50.02 & 48.06 & \multicolumn{1}{c|}{56.09} & 55.55 & 52.08 & 59.52 \\
\multicolumn{1}{c|}{FedAWA} & \multicolumn{1}{c|}{CVPR'25} & 65.58 & 60.72 & \multicolumn{1}{c|}{61.20} & 62.09 & \underline{59.73} & \multicolumn{1}{c|}{60.50} & 61.47 & 57.58 & \multicolumn{1}{c|}{59.14} & 60.14 & 56.67 & \multicolumn{1}{c|}{59.46} & 59.01 & 52.69 & \multicolumn{1}{c|}{58.77} & 57.57 & 49.65 & \multicolumn{1}{c|}{57.73} & 60.98 & 56.17 & 59.47 \\
\multicolumn{1}{c|}{Fed-RAM} & \multicolumn{1}{c|}{CVPR'25} & \underline{66.06} & 60.98 & \multicolumn{1}{c|}{62.89} & 63.08 & 57.97 & \multicolumn{1}{c|}{64.45} & 59.97 & \underline{57.97} & \multicolumn{1}{c|}{\underline{64.45}} & \underline{60.85} & 56.95 & \multicolumn{1}{c|}{\underline{62.73}} & \underline{60.88} & \underline{53.00} & \multicolumn{1}{c|}{56.25} & 55.64 & 50.37 & \multicolumn{1}{c|}{55.04} & 61.08 & 56.21 & 60.97 \\
\multicolumn{1}{c|}{FedMVP} & \multicolumn{1}{c|}{ICCV'25} & 65.76 & 58.06 & \multicolumn{1}{c|}{\underline{64.01}} & \underline{63.19} & 56.75 & \multicolumn{1}{c|}{\underline{65.48}} & \underline{62.26} & 57.20 & \multicolumn{1}{c|}{63.51} & 60.60 & 53.42 & \multicolumn{1}{c|}{62.66} & 59.56 & 51.84 & \multicolumn{1}{c|}{59.51} & \underline{58.48} & \underline{50.42} & \multicolumn{1}{c|}{\underline{55.30}} & \underline{61.64} & 54.62 & \underline{61.75} \\
\rowcolor{gray!10}\multicolumn{1}{c|}{FedMPT} & \multicolumn{1}{c|}{Ours} & \textbf{67.37} & \textbf{62.88} & \multicolumn{1}{c|}{\textbf{67.57}} & \textbf{65.91} & \textbf{60.38} & \multicolumn{1}{c|}{\textbf{67.11}} & \textbf{65.31} & \textbf{59.10} & \multicolumn{1}{c|}{\textbf{66.07}} & \textbf{63.86} & \textbf{58.46} & \multicolumn{1}{c|}{\textbf{64.32}} & \textbf{63.09} & \textbf{57.19} & \multicolumn{1}{c|}{\textbf{63.68}} & \textbf{62.37} & \textbf{56.96} & \multicolumn{1}{c|}{\textbf{62.80}} & \textbf{64.65} & \textbf{59.16} & \textbf{65.26} \\
\rowcolor{gray!10}\multicolumn{1}{c|}{\textbf{$\Delta$ Prev. best}} & \multicolumn{1}{c|}{\textbackslash{}} & \textcolor{red}{\textbf{+1.31}} & \textcolor{red}{\textbf{+1.38}} & \multicolumn{1}{c|}{\textcolor{red}{\textbf{+0.41}}} & \textcolor{red}{\textbf{+2.72}} & \textcolor{red}{\textbf{+0.65}} & \multicolumn{1}{c|}{\textcolor{red}{\textbf{+1.63}}} & \textcolor{red}{\textbf{+3.05}} & \textcolor{red}{\textbf{+0.85}} & \multicolumn{1}{c|}{\textcolor{red}{\textbf{+1.62}}} & \textcolor{red}{\textbf{+3.01}} & \textcolor{red}{\textbf{+0.63}} & \multicolumn{1}{c|}{\textcolor{red}{\textbf{+1.59}}} & \textcolor{red}{\textbf{+2.21}} & \textcolor{red}{\textbf{+4.19}} & \multicolumn{1}{c|}{\textcolor{red}{\textbf{+2.83}}} & \textcolor{red}{\textbf{+3.89}} & \textcolor{red}{\textbf{+3.77}} & \multicolumn{1}{c|}{\textcolor{red}{\textbf{+5.07}}} & \textcolor{red}{\textbf{+3.01}} & \textcolor{red}{\textbf{+2.78}} & \textcolor{red}{\textbf{+3.51}} \\ \hline
\multicolumn{23}{c}{NUS-Wide} \\ \hline
\multicolumn{1}{c|}{} & \multicolumn{1}{c|}{} & \multicolumn{3}{c|}{$t=10\%$} & \multicolumn{3}{c|}{$t=20\%$} & \multicolumn{3}{c|}{$t=40\%$} & \multicolumn{3}{c|}{$t=60\%$} & \multicolumn{3}{c|}{$t=80\%$} & \multicolumn{3}{c|}{$t=100\%$} & \multicolumn{3}{c}{Avg} \\ \cline{3-23} 
\multicolumn{1}{c|}{\multirow{-2}{*}{Methods}} & \multicolumn{1}{c|}{\multirow{-2}{*}{Venue}} & mAP & CF1 & \multicolumn{1}{c|}{OF1} & mAP & CF1 & \multicolumn{1}{c|}{OF1} & mAP & CF1 & \multicolumn{1}{c|}{OF1} & mAP & CF1 & \multicolumn{1}{c|}{OF1} & mAP & CF1 & \multicolumn{1}{c|}{OF1} & mAP & CF1 & \multicolumn{1}{c|}{OF1} & mAP & CF1 & OF1 \\ \hline
\multicolumn{1}{c|}{Fed-DualCoOp} & \multicolumn{1}{c|}{NeurIPS'22} & 51.02 & 45.89 & \multicolumn{1}{c|}{69.00} & 49.97 & 44.36 & \multicolumn{1}{c|}{65.30} & 46.25 & 41.62 & \multicolumn{1}{c|}{67.02} & 42.69 & 37.11 & \multicolumn{1}{c|}{66.10} & 40.65 & 35.69 & \multicolumn{1}{c|}{60.28} & 39.40 & 34.54 & \multicolumn{1}{c|}{59.04} & 45.00 & 39.87 & 64.46 \\
\multicolumn{1}{c|}{Fed-SCPNet} & \multicolumn{1}{c|}{CVPR'23} & 49.56 & 42.11 & \multicolumn{1}{c|}{65.63} & 47.61 & 39.98 & \multicolumn{1}{c|}{59.46} & 43.31 & 40.92 & \multicolumn{1}{c|}{62.04} & 39.28 & 36.51 & \multicolumn{1}{c|}{60.45} & 38.61 & 39.98 & \multicolumn{1}{c|}{58.46} & 38.49 & 39.02 & \multicolumn{1}{c|}{53.23} & 42.81 & 39.75 & 59.88 \\
\multicolumn{1}{c|}{Fed-MaPLe} & \multicolumn{1}{c|}{CVPR'23} & 54.78 & 46.69 & \multicolumn{1}{c|}{72.32} & 53.68 & 42.13 & \multicolumn{1}{c|}{71.12} & 51.34 & 40.69 & \multicolumn{1}{c|}{73.32} & 51.62 & 43.40 & \multicolumn{1}{c|}{72.53} & 50.98 & 43.68 & \multicolumn{1}{c|}{72.70} & 51.20 & 40.99 & \multicolumn{1}{c|}{70.59} & 52.27 & 42.93 & 72.10 \\
\multicolumn{1}{c|}{FedPGP} & \multicolumn{1}{c|}{ICML'24} & 55.92 & 46.48 & \multicolumn{1}{c|}{75.91} & 53.41 & 43.27 & \multicolumn{1}{c|}{75.28} & 53.78 & 43.07 & \multicolumn{1}{c|}{72.56} & 49.64 & 44.82 & \multicolumn{1}{c|}{72.79} & 48.45 & 42.58 & \multicolumn{1}{c|}{72.47} & 43.20 & 42.72 & \multicolumn{1}{c|}{71.73} & 50.73 & 43.82 & 73.46 \\
\multicolumn{1}{c|}{Fed-TCP} & \multicolumn{1}{c|}{CVPR'24} & 54.46 & 48.23 & \multicolumn{1}{c|}{70.42} & 51.93 & 46.57 & \multicolumn{1}{c|}{70.20} & 50.91 & 45.65 & \multicolumn{1}{c|}{69.96} & 48.28 & 42.47 & \multicolumn{1}{c|}{65.44} & 47.20 & 41.75 & \multicolumn{1}{c|}{66.80} & 48.89 & 40.50 & \multicolumn{1}{c|}{63.17} & 50.28 & 44.20 & 67.67 \\
\multicolumn{1}{c|}{FedTPG} & \multicolumn{1}{c|}{ICLR'24} & 53.01 & \underline{50.69} & \multicolumn{1}{c|}{73.32} & 52.78 & 47.66 & \multicolumn{1}{c|}{72.50} & 50.00 & 46.72 & \multicolumn{1}{c|}{69.52} & 50.40 & 44.21 & \multicolumn{1}{c|}{70.47} & 49.92 & 43.89 & \multicolumn{1}{c|}{67.10} & 49.46 & 42.55 & \multicolumn{1}{c|}{65.46} & 50.93 & 45.95 & 69.73 \\
\multicolumn{1}{c|}{Fed-PosCoOp} & \multicolumn{1}{c|}{WACV'25} & 52.88 & 47.96 & \multicolumn{1}{c|}{70.38} & 50.64 & 45.89 & \multicolumn{1}{c|}{70.20} & 48.77 & 42.69 & \multicolumn{1}{c|}{68.66} & 42.39 & 39.15 & \multicolumn{1}{c|}{68.04} & 40.62 & 35.12 & \multicolumn{1}{c|}{64.33} & 40.08 & 32.76 & \multicolumn{1}{c|}{60.11} & 45.90 & 40.60 & 66.95 \\
\multicolumn{1}{c|}{FedAWA} & \multicolumn{1}{c|}{CVPR'25} & 54.67 & 47.61 & \multicolumn{1}{c|}{74.51} & 53.91 & 45.52 & \multicolumn{1}{c|}{70.33} & 50.40 & 43.74 & \multicolumn{1}{c|}{71.60} & 52.87 & 43.22 & \multicolumn{1}{c|}{71.51} & 52.75 & 43.02 & \multicolumn{1}{c|}{66.89} & 50.71 & 41.13 & \multicolumn{1}{c|}{66.53} & 52.55 & 44.04 & 70.23 \\
\multicolumn{1}{c|}{Fed-RAM} & \multicolumn{1}{c|}{CVPR'25} & \underline{56.35} & 48.89 & \multicolumn{1}{c|}{76.07} & \underline{54.05} & \underline{47.89} & \multicolumn{1}{c|}{76.40} & \underline{53.51} & \underline{45.69} & \multicolumn{1}{c|}{74.33} & 52.25 & 40.69 & \multicolumn{1}{c|}{74.38} & \underline{52.61} & 42.12 & \multicolumn{1}{c|}{72.98} & \underline{51.20} & \underline{40.69} & \multicolumn{1}{c|}{70.35} & \underline{53.33} & 44.33 & 74.09 \\
\multicolumn{1}{c|}{FedMVP} & \multicolumn{1}{c|}{ICCV'25} & 53.20 & 48.37 & \multicolumn{1}{c|}{\underline{77.80}} & 53.32 & 46.39 & \multicolumn{1}{c|}{\underline{77.10}} & 51.90 & 45.21 & \multicolumn{1}{c|}{\underline{75.14}} & \underline{52.98} & \underline{43.43} & \multicolumn{1}{c|}{\underline{75.02}} & 51.33 & \underline{42.79} & \multicolumn{1}{c|}{\underline{74.55}} & 51.08 & 40.16 & \multicolumn{1}{c|}{\underline{72.90}} & 52.30 & \underline{44.39} & \underline{75.42} \\
\rowcolor{gray!10}\multicolumn{1}{c|}{FedMPT} & \multicolumn{1}{c|}{Ours} & \textbf{58.36} & \textbf{51.56} & \multicolumn{1}{c|}{\textbf{78.98}} & \textbf{57.27} & \textbf{49.29} & \multicolumn{1}{c|}{\textbf{77.93}} & \textbf{57.03} & \textbf{48.14} & \multicolumn{1}{c|}{\textbf{76.73}} & \textbf{56.52} & \textbf{46.88} & \multicolumn{1}{c|}{\textbf{76.94}} & \textbf{56.08} & \textbf{45.73} & \multicolumn{1}{c|}{\textbf{76.60}} & \textbf{54.87} & \textbf{45.03} & \multicolumn{1}{c|}{\textbf{76.81}} & \textbf{56.69} & \textbf{47.77} & \textbf{77.33} \\
\rowcolor{gray!10}\multicolumn{1}{c|}{\textbf{$\Delta$ Prev. best}} & \multicolumn{1}{c|}{\textbackslash{}} & \textcolor{red}{\textbf{+2.01}} & \textcolor{red}{\textbf{+0.87}} & \multicolumn{1}{c|}{\textcolor{red}{\textbf{+1.18}}} & \textcolor{red}{\textbf{+3.22}} & \textcolor{red}{\textbf{+1.40}} & \multicolumn{1}{c|}{\textcolor{red}{\textbf{+0.83}}} & \textcolor{red}{\textbf{+3.25}} & \textcolor{red}{\textbf{+1.42}} & \multicolumn{1}{c|}{\textcolor{red}{\textbf{+1.59}}} & \textcolor{red}{\textbf{+3.54}} & \textcolor{red}{\textbf{+2.06}} & \multicolumn{1}{c|}{\textcolor{red}{\textbf{+1.92}}} & \textcolor{red}{\textbf{+3.33}} & \textcolor{red}{\textbf{+1.84}} & \multicolumn{1}{c|}{\textcolor{red}{\textbf{+2.05}}} & \textcolor{red}{\textbf{+3.67}} & \textcolor{red}{\textbf{+2.31}} & \multicolumn{1}{c|}{\textcolor{red}{\textbf{+3.91}}} & \textcolor{red}{\textbf{+3.36}} & \textcolor{red}{\textbf{+1.82}} & \textcolor{red}{\textbf{+1.91}} \\ \hline
\end{tabular}
} 
\end{center}\label{tab:11}
\vspace{-0.8cm}
\end{table*}

\section{Experiments}
\paragraph{Benchmarks.} We systematically evaluate the effectiveness of FedMPT under the federated MLR setting with three benchmarks: \underline{\ding{182} Heterogeneity Benchmark}, which assesses model robustness to varying degrees of data heterogeneity across clients. To achieve this perspective, the training dataset is first partitioned into $S$ clusters according to their visual embeddings from ViT/B-16, then each data cluster is assigned to a client. We change $S$ by varying $t(\%)$, the proportion of the class size $C$.  \underline{\ding{183} Federated Part-Annotation  Benchmark}, which evaluates the models' robustness to insufficient annotations by randomly masking `\texttt{Mask}' class annotations in the training set. The heterogeneity setting $t$ is kept as $60\%$ for this benchmark. We employ three datasets for the above benchmarks: VOC2007~\cite{everingham2015pascal}, COCO2014~\cite{lin2014microsoft}, and NUS-wide~\cite{chua2009nus}. Furthermore, to assess models' real-world applicability, we adopt \underline{\ding{184} Federated Real-world MLR Benchmark}\cite{tan2025recover}, which incorporates two remote sensing datasets: Multi-Sense~\cite{hua2021multiscene} and MLRSNet~\cite{qi2020mlrsnet}. $t=60\%$. \textit{We evaluate the performance of the global model on the test sets. (For methods~\cite{cui2024harmonizing} that also maintain local private parameters in clients, we evaluate the performance of each client on the test sets and average them).} Following~\cite{ma2025correlative}, we report the three most important metrics in all benchmarks: Mean Average Precision (mAP), per-category F1-score (CF1), and overall F1-score (OF1). All reported results are the average of 3 independent runs. Experiments on other datasets, finer settings, or benchmarks (like ZSL~\cite{sun2022dualcoop}) are in \textbf{\underline{Sup. Mat. A.}}

We select the following competitive baselines spanning MLR, FL and Prompt Learning for comprehensive comparisons: DualCoOp~\cite{sun2022dualcoop}, SCPNet~\cite{ding2023exploring}, PosCoOp~\cite{rawlekar2025positivecoop} and RAM~\cite{tan2025recover}, which are competitive baselines of MLR with VLMs; MaPLe~\cite{khattak2023maple}\&TCP~\cite{yao2024tcp}, which are typical representations of prompt learning with VLMs; FedPGP~\cite{cui2024harmonizing}, FedTPG~\cite{qiu2024federated}, FedAWA~\cite{shi2025fedawa} (For the fairness of comparison, we apply FedAWA to the prompt learner of CoOp~\cite{zhou2022learning}) and FedMVP~\cite{singha2025fedmvp}, which are state-of-the-arts of federated learning with VLMs. Notably, for methods that are not originally designed for federated scenarios, we alter their methodology by training a local model for each client (with their original cross-entropy loss altered to $\mathcal{L}_{asl}$ mentioned in Sec \ref{sec31}) and aggregating the weights of their learnable modules with FedAvg. We add a ``Fed-'' prefix to the names of these methods to highlight our modification. More details of datasets and baselines are in \textbf{\underline{Sup. Mat. C.}}.

\begin{table*}[t]
\caption{Comparison of FedMPT and other methods on the Federated Part-Annotation Benchmark. We report the mAP, CF1 and OF1 with the part-annotation setting \texttt{Mask} varying from 10\% to 90\%. The best results are marked with \textbf{bold}.}
\vspace{-0.5cm}
\begin{center} 
\renewcommand{\arraystretch}{1}   
\setlength\tabcolsep{5pt} 
\resizebox{1\textwidth}{!}{
\begin{tabular}{cccccccccccccccccccc}
\hline
\multicolumn{20}{c}{VOC2007} \\ \hline
\multicolumn{1}{c|}{} & \multicolumn{1}{c|}{} & \multicolumn{3}{c|}{$\texttt{Mask}=10\%$} & \multicolumn{3}{c|}{$\texttt{Mask}=30\%$} & \multicolumn{3}{c|}{$\texttt{Mask}=50\%$} & \multicolumn{3}{c|}{$\texttt{Mask}=70\%$} & \multicolumn{3}{c|}{$\texttt{Mask}=90\%$} & \multicolumn{3}{c}{Avg} \\ \cline{3-20} 
\multicolumn{1}{c|}{\multirow{-2}{*}{Methods}} & \multicolumn{1}{c|}{\multirow{-2}{*}{Venue}} & mAP & CF1 & \multicolumn{1}{c|}{OF1} & mAP & CF1 & \multicolumn{1}{c|}{OF1} & mAP & CF1 & \multicolumn{1}{c|}{OF1} & mAP & CF1 & \multicolumn{1}{c|}{OF1} & mAP & CF1 & \multicolumn{1}{c|}{OF1} & mAP & CF1 & OF1 \\ \hline
\multicolumn{1}{c|}{Fed-DualCoOp} & \multicolumn{1}{c|}{NeurIPS'22} & 82.52 & 76.86 & \multicolumn{1}{c|}{76.58} & 80.52 & 75.00 & \multicolumn{1}{c|}{74.73} & 77.65 & 72.65 & \multicolumn{1}{c|}{69.98} & 74.43 & 66.72 & \multicolumn{1}{c|}{63.39} & 61.28 & 53.89 & \multicolumn{1}{c|}{50.96} & 75.28 & 69.02 & 67.13 \\
\multicolumn{1}{c|}{Fed-SCPNet} & \multicolumn{1}{c|}{CVPR'23} & 78.81 & 70.83 & \multicolumn{1}{c|}{72.60} & 73.55 & 72.04 & \multicolumn{1}{c|}{65.77} & 70.97 & 70.06 & \multicolumn{1}{c|}{70.62} & 67.46 & 60.09 & \multicolumn{1}{c|}{55.04} & 64.73 & 56.51 & \multicolumn{1}{c|}{52.93} & 71.10 & 65.91 & 63.39 \\
\multicolumn{1}{c|}{Fed-MaPLe} & \multicolumn{1}{c|}{CVPR'23} & 81.27 & 75.59 & \multicolumn{1}{c|}{73.49} & 81.16 & 75.46 & \multicolumn{1}{c|}{74.11} & 78.50 & 76.30 & \multicolumn{1}{c|}{\underline{74.11}} & 70.60 & 65.60 & \multicolumn{1}{c|}{64.75} & 59.53 & 58.64 & \multicolumn{1}{c|}{48.62} & 74.21 & 70.32 & 67.02 \\
\multicolumn{1}{c|}{FedPGP} & \multicolumn{1}{c|}{ICML'24} & 79.96 & 73.81 & \multicolumn{1}{c|}{63.81} & 76.71 & 66.44 & \multicolumn{1}{c|}{64.05} & 72.96 & 67.81 & \multicolumn{1}{c|}{63.81} & 68.29 & 66.27 & \multicolumn{1}{c|}{59.70} & 66.05 & 60.54 & \multicolumn{1}{c|}{50.65} & 72.79 & 66.97 & 60.40 \\
\multicolumn{1}{c|}{Fed-TCP} & \multicolumn{1}{c|}{CVPR'24} & 76.42 & 76.00 & \multicolumn{1}{c|}{72.50} & 75.42 & 74.09 & \multicolumn{1}{c|}{72.54} & 72.43 & 70.03 & \multicolumn{1}{c|}{70.59} & 70.35 & 68.90 & \multicolumn{1}{c|}{67.86} & 67.51 & 62.80 & \multicolumn{1}{c|}{55.90} & 72.43 & 70.36 & 67.88 \\
\multicolumn{1}{c|}{FedTPG} & \multicolumn{1}{c|}{ICLR'24} & 84.19 & 77.45 & \multicolumn{1}{c|}{\underline{76.93}} & 82.68 & 75.31 & \multicolumn{1}{c|}{67.00} & 82.18 & 68.19 & \multicolumn{1}{c|}{56.87} & 80.00 & 65.85 & \multicolumn{1}{c|}{59.29} & \underline{68.24} & 58.68 & \multicolumn{1}{c|}{57.09} & 79.46 & 69.10 & 63.44 \\
\multicolumn{1}{c|}{Fed-PosCoOp} & \multicolumn{1}{c|}{WACV'25} & 81.18 & 76.44 & \multicolumn{1}{c|}{73.29} & 79.86 & 76.13 & \multicolumn{1}{c|}{72.96} & 78.50 & 64.62 & \multicolumn{1}{c|}{71.73} & 73.87 & 65.63 & \multicolumn{1}{c|}{67.74} & 60.27 & 58.90 & \multicolumn{1}{c|}{54.65} & 74.74 & 68.34 & 68.07 \\
\multicolumn{1}{c|}{FedAWA} & \multicolumn{1}{c|}{CVPR'25} & 80.81 & 70.70 & \multicolumn{1}{c|}{72.94} & 75.37 & 69.16 & \multicolumn{1}{c|}{65.68} & 75.82 & 69.66 & \multicolumn{1}{c|}{65.87} & 72.59 & 67.45 & \multicolumn{1}{c|}{57.16} & 68.22 & 55.27 & \multicolumn{1}{c|}{56.92} & 74.56 & 66.45 & 63.71 \\
\multicolumn{1}{c|}{Fed-RAM} & \multicolumn{1}{c|}{CVPR'25} & \underline{87.89} & 77.03 & \multicolumn{1}{c|}{75.93} & \underline{85.42} & 75.78 & \multicolumn{1}{c|}{73.41} & \underline{82.30} & \underline{74.84} & \multicolumn{1}{c|}{72.93} & 79.58 & 72.67 & \multicolumn{1}{c|}{68.41} & 68.09 & \underline{63.18} & \multicolumn{1}{c|}{\underline{59.84}} & \underline{80.66} & 72.70 & 70.10 \\
\multicolumn{1}{c|}{FedMVP} & \multicolumn{1}{c|}{ICCV'25} & 85.27 & \underline{79.10} & \multicolumn{1}{c|}{76.23} & 84.34 & \underline{76.83} & \multicolumn{1}{c|}{\underline{75.99}} & 82.09 & 74.20 & \multicolumn{1}{c|}{\underline{73.07}} & \underline{80.64} & \underline{75.03} & \multicolumn{1}{c|}{\underline{70.15}} & 66.53 & 62.89 & \multicolumn{1}{c|}{58.40} & 79.77 & \underline{73.61} & \underline{70.77} \\
\rowcolor{gray!10}\multicolumn{1}{c|}{FedMPT} & \multicolumn{1}{c|}{Ours} & \textbf{89.40} & \textbf{83.06} & \multicolumn{1}{c|}{\textbf{81.12}} & \textbf{87.40} & \textbf{82.12} & \multicolumn{1}{c|}{\textbf{81.48}} & \textbf{88.75} & \textbf{80.86} & \multicolumn{1}{c|}{\textbf{79.71}} & \textbf{86.18} & \textbf{81.32} & \multicolumn{1}{c|}{\textbf{74.47}} & \textbf{74.24} & \textbf{66.23} & \multicolumn{1}{c|}{\textbf{64.93}} & \textbf{85.19} & \textbf{78.72} & \textbf{76.34} \\
\rowcolor{gray!10}\multicolumn{1}{c|}{\textbf{$\Delta$ Prev. Best}} & \multicolumn{1}{c|}{\textbackslash{}} & \textcolor{red}{\textbf{+1.51}} & \textcolor{red}{\textbf{+3.96}} & \multicolumn{1}{c|}{\textcolor{red}{\textbf{+4.19}}} & \textcolor{red}{\textbf{+1.98}} & \textcolor{red}{\textbf{+5.29}} & \multicolumn{1}{c|}{\textcolor{red}{\textbf{+5.49}}} & \textcolor{red}{\textbf{+6.45}} & \textcolor{red}{\textbf{+4.56}} & \multicolumn{1}{c|}{\textcolor{red}{\textbf{+5.60}}} & \textcolor{red}{\textbf{+5.54}} & \textcolor{red}{\textbf{+6.29}} & \multicolumn{1}{c|}{\textcolor{red}{\textbf{+4.32}}} & \textcolor{red}{\textbf{+6.00}} & \textcolor{red}{\textbf{+3.05}} & \multicolumn{1}{c|}{\textcolor{red}{\textbf{+5.09}}} & \textcolor{red}{\textbf{+4.54}} & \textcolor{red}{\textbf{+5.11}} & \textcolor{red}{\textbf{+5.57}} \\ \hline
\multicolumn{20}{c}{COCO2014} \\ \hline
\multicolumn{1}{c|}{} & \multicolumn{1}{c|}{} & \multicolumn{3}{c|}{$\texttt{Mask}=10\%$} & \multicolumn{3}{c|}{$\texttt{Mask}=30\%$} & \multicolumn{3}{c|}{$\texttt{Mask}=50\%$} & \multicolumn{3}{c|}{$\texttt{Mask}=70\%$} & \multicolumn{3}{c|}{$\texttt{Mask}=90\%$} & \multicolumn{3}{c}{Avg} \\ \cline{3-20} 
\multicolumn{1}{c|}{\multirow{-2}{*}{Methods}} & \multicolumn{1}{c|}{\multirow{-2}{*}{Venue}} & mAP & CF1 & \multicolumn{1}{c|}{OF1} & mAP & CF1 & \multicolumn{1}{c|}{OF1} & mAP & CF1 & \multicolumn{1}{c|}{OF1} & mAP & CF1 & \multicolumn{1}{c|}{OF1} & mAP & CF1 & \multicolumn{1}{c|}{OF1} & mAP & CF1 & OF1 \\ \hline
\multicolumn{1}{c|}{Fed-DualCoOp} & \multicolumn{1}{c|}{NeurIPS'22} & 52.44 & 47.09 & \multicolumn{1}{c|}{56.86} & 39.56 & 42.05 & \multicolumn{1}{c|}{37.50} & 26.55 & 28.57 & \multicolumn{1}{c|}{22.04} & 22.72 & 24.40 & \multicolumn{1}{c|}{21.50} & 20.96 & 22.97 & \multicolumn{1}{c|}{12.76} & 32.45 & 33.02 & 30.13 \\
\multicolumn{1}{c|}{Fed-SCPNet} & \multicolumn{1}{c|}{CVPR'23} & 50.11 & 46.30 & \multicolumn{1}{c|}{55.80} & 38.75 & 35.24 & \multicolumn{1}{c|}{33.56} & 23.08 & 26.73 & \multicolumn{1}{c|}{23.71} & 24.46 & 26.82 & \multicolumn{1}{c|}{20.74} & 21.15 & 14.82 & \multicolumn{1}{c|}{13.70} & 31.51 & 29.98 & 29.50 \\
\multicolumn{1}{c|}{Fed-MaPLe} & \multicolumn{1}{c|}{CVPR'23} & 57.95 & 54.96 & \multicolumn{1}{c|}{56.32} & 36.86 & 39.54 & \multicolumn{1}{c|}{30.13} & 25.16 & 27.78 & \multicolumn{1}{c|}{25.22} & 22.61 & 24.36 & \multicolumn{1}{c|}{22.82} & 20.86 & 18.49 & \multicolumn{1}{c|}{17.23} & 32.69 & 33.03 & 30.34 \\
\multicolumn{1}{c|}{FedPGP} & \multicolumn{1}{c|}{ICML'24} & 58.14 & 51.23 & \multicolumn{1}{c|}{54.25} & 35.30 & 38.89 & \multicolumn{1}{c|}{38.32} & 23.12 & 29.44 & \multicolumn{1}{c|}{20.36} & 21.57 & 28.09 & \multicolumn{1}{c|}{20.84} & 23.14 & 16.19 & \multicolumn{1}{c|}{17.15} & 32.25 & 32.77 & 30.18 \\
\multicolumn{1}{c|}{Fed-TCP} & \multicolumn{1}{c|}{CVPR'24} & 57.88 & 50.20 & \multicolumn{1}{c|}{55.15} & 40.70 & 41.85 & \multicolumn{1}{c|}{25.07} & 24.61 & 27.73 & \multicolumn{1}{c|}{30.54} & 22.57 & 24.51 & \multicolumn{1}{c|}{19.57} & 21.00 & 18.48 & \multicolumn{1}{c|}{12.57} & 33.35 & 32.55 & 28.58 \\
\multicolumn{1}{c|}{FedTPG} & \multicolumn{1}{c|}{ICLR'24} & 60.49 & 52.85 & \multicolumn{1}{c|}{53.17} & 42.71 & 40.08 & \multicolumn{1}{c|}{41.60} & 21.40 & 25.88 & \multicolumn{1}{c|}{26.74} & 23.28 & 25.32 & \multicolumn{1}{c|}{22.97} & 20.93 & 16.55 & \multicolumn{1}{c|}{14.65} & 33.76 & 32.14 & 31.83 \\
\multicolumn{1}{c|}{Fed-PosCoOp} & \multicolumn{1}{c|}{WACV'25} & 52.18 & 49.70 & \multicolumn{1}{c|}{54.06} & 39.70 & 40.04 & \multicolumn{1}{c|}{40.58} & 27.03 & 23.68 & \multicolumn{1}{c|}{27.23} & 22.49 & 24.98 & \multicolumn{1}{c|}{23.67} & 19.77 & 13.00 & \multicolumn{1}{c|}{19.57} & 32.23 & 30.28 & 33.02 \\
\multicolumn{1}{c|}{FedAWA} & \multicolumn{1}{c|}{CVPR'25} & 58.80 & 52.55 & \multicolumn{1}{c|}{58.94} & 38.40 & 38.06 & \multicolumn{1}{c|}{39.30} & 28.06 & 28.80 & \multicolumn{1}{c|}{34.64} & 21.00 & 26.81 & \multicolumn{1}{c|}{24.50} & 18.37 & 18.16 & \multicolumn{1}{c|}{23.13} & 32.93 & 32.88 & 36.10 \\
\multicolumn{1}{c|}{Fed-RAM} & \multicolumn{1}{c|}{CVPR'25} & \underline{60.80} & \underline{55.85} & \multicolumn{1}{c|}{\underline{59.89}} & \underline{42.68} & \underline{42.82} & \multicolumn{1}{c|}{\underline{50.34}} & \underline{29.79} & \underline{33.25} & \multicolumn{1}{c|}{\underline{38.51}} & 24.86 & 27.63 & \multicolumn{1}{c|}{\underline{26.42}} & 20.26 & 19.52 & \multicolumn{1}{c|}{23.50} & 35.68 & \underline{35.81} & \underline{39.73} \\
\multicolumn{1}{c|}{FedMVP} & \multicolumn{1}{c|}{ICCV'25} & 58.80 & 53.75 & \multicolumn{1}{c|}{55.10} & 41.64 & 39.51 & \multicolumn{1}{c|}{45.52} & 28.46 & 28.19 & \multicolumn{1}{c|}{36.07} & \underline{26.26} & \underline{29.82} & \multicolumn{1}{c|}{25.53} & \underline{25.82} & \underline{24.86} & \multicolumn{1}{c|}{\underline{24.54}} & \underline{36.20} & 35.23 & 37.35 \\
\rowcolor{gray!10}\multicolumn{1}{c|}{FedMPT} & \multicolumn{1}{c|}{Ours} & \textbf{62.75} & \textbf{55.29} & \multicolumn{1}{c|}{\textbf{61.02}} & \textbf{45.98} & \textbf{44.68} & \multicolumn{1}{c|}{\textbf{54.25}} & \textbf{33.13} & \textbf{36.22} & \multicolumn{1}{c|}{\textbf{44.22}} & \textbf{30.54} & \textbf{33.95} & \multicolumn{1}{c|}{\textbf{32.06}} & \textbf{30.39} & \textbf{32.84} & \multicolumn{1}{c|}{\textbf{30.78}} & \textbf{40.56} & \textbf{40.60} & \textbf{44.47} \\
\rowcolor{gray!10}\multicolumn{1}{c|}{\textbf{$\Delta$ Prev. Best}} & \multicolumn{1}{c|}{\textbackslash{}} & \textcolor{red}{\textbf{+1.95}} & \textcolor{red}{\textbf{+-0.56}} & \multicolumn{1}{c|}{\textcolor{red}{\textbf{+1.13}}} & \textcolor{red}{\textbf{+3.27}} & \textcolor{red}{\textbf{+1.86}} & \multicolumn{1}{c|}{\textcolor{red}{\textbf{+3.91}}} & \textcolor{red}{\textbf{+3.34}} & \textcolor{red}{\textbf{+2.97}} & \multicolumn{1}{c|}{\textcolor{red}{\textbf{+5.71}}} & \textcolor{red}{\textbf{+4.28}} & \textcolor{red}{\textbf{+4.13}} & \multicolumn{1}{c|}{\textcolor{red}{\textbf{+5.64}}} & \textcolor{red}{\textbf{+4.57}} & \textcolor{red}{\textbf{+7.98}} & \multicolumn{1}{c|}{\textcolor{red}{\textbf{+6.24}}} & \textcolor{red}{\textbf{+4.36}} & \textcolor{red}{\textbf{+4.78}} & \textcolor{red}{\textbf{+4.73}} \\ \hline
\multicolumn{20}{c}{NUS-Wide} \\ \hline
\multicolumn{1}{c|}{} & \multicolumn{1}{c|}{} & \multicolumn{3}{c|}{$\texttt{Mask}=10\%$} & \multicolumn{3}{c|}{$\texttt{Mask}=30\%$} & \multicolumn{3}{c|}{$\texttt{Mask}=50\%$} & \multicolumn{3}{c|}{$\texttt{Mask}=70\%$} & \multicolumn{3}{c|}{$\texttt{Mask}=90\%$} & \multicolumn{3}{c}{Avg} \\ \cline{3-20} 
\multicolumn{1}{c|}{\multirow{-2}{*}{Methods}} & \multicolumn{1}{c|}{\multirow{-2}{*}{Venue}} & mAP & CF1 & \multicolumn{1}{c|}{OF1} & mAP & CF1 & \multicolumn{1}{c|}{OF1} & mAP & CF1 & \multicolumn{1}{c|}{OF1} & mAP & CF1 & \multicolumn{1}{c|}{OF1} & mAP & CF1 & \multicolumn{1}{c|}{OF1} & mAP & CF1 & OF1 \\ \hline
\multicolumn{1}{c|}{Fed-DualCoOp} & \multicolumn{1}{c|}{NeurIPS'22} & 40.13 & 35.57 & \multicolumn{1}{c|}{61.16} & 31.20 & 22.90 & \multicolumn{1}{c|}{43.49} & 24.79 & 23.03 & \multicolumn{1}{c|}{28.87} & 13.04 & 9.04 & \multicolumn{1}{c|}{16.05} & 10.81 & 9.34 & \multicolumn{1}{c|}{4.17} & 23.99 & 19.98 & 30.75 \\
\multicolumn{1}{c|}{Fed-SCPNet} & \multicolumn{1}{c|}{CVPR'23} & 35.44 & 33.22 & \multicolumn{1}{c|}{56.39} & 28.95 & 20.45 & \multicolumn{1}{c|}{39.91} & 23.64 & 21.99 & \multicolumn{1}{c|}{26.73} & 13.28 & 8.96 & \multicolumn{1}{c|}{19.90} & 9.97 & 9.61 & \multicolumn{1}{c|}{6.24} & 22.26 & 18.85 & 29.83 \\
\multicolumn{1}{c|}{Fed-MaPLe} & \multicolumn{1}{c|}{CVPR'23} & 48.60 & 40.67 & \multicolumn{1}{c|}{71.28} & 30.20 & 25.62 & \multicolumn{1}{c|}{50.50} & 25.10 & 22.49 & \multicolumn{1}{c|}{37.35} & 13.85 & 11.06 & \multicolumn{1}{c|}{26.01} & 6.28 & 5.53 & \multicolumn{1}{c|}{11.20} & 24.81 & 21.07 & 39.27 \\
\multicolumn{1}{c|}{FedPGP} & \multicolumn{1}{c|}{ICML'24} & 42.24 & 36.90 & \multicolumn{1}{c|}{68.55} & 36.54 & 32.36 & \multicolumn{1}{c|}{55.74} & 23.24 & 26.90 & \multicolumn{1}{c|}{41.55} & 19.67 & 12.60 & \multicolumn{1}{c|}{28.03} & 7.98 & 4.95 & \multicolumn{1}{c|}{9.54} & 25.93 & 22.74 & 40.68 \\
\multicolumn{1}{c|}{Fed-TCP} & \multicolumn{1}{c|}{CVPR'24} & 47.19 & 40.70 & \multicolumn{1}{c|}{61.81} & 35.40 & 35.53 & \multicolumn{1}{c|}{43.67} & 26.86 & 26.58 & \multicolumn{1}{c|}{34.40} & 12.38 & 12.06 & \multicolumn{1}{c|}{15.34} & 9.86 & 9.46 & \multicolumn{1}{c|}{12.66} & 26.34 & 24.87 & 33.58 \\
\multicolumn{1}{c|}{FedTPG} & \multicolumn{1}{c|}{ICLR'24} & 47.64 & 38.60 & \multicolumn{1}{c|}{66.39} & 34.60 & 32.17 & \multicolumn{1}{c|}{52.85} & 26.85 & 28.76 & \multicolumn{1}{c|}{41.05} & 20.53 & 12.23 & \multicolumn{1}{c|}{22.28} & 11.52 & 10.47 & \multicolumn{1}{c|}{12.89} & 28.23 & 24.45 & 39.09 \\
\multicolumn{1}{c|}{Fed-PosCoOp} & \multicolumn{1}{c|}{WACV'25} & 40.64 & 39.38 & \multicolumn{1}{c|}{63.42} & 35.94 & 34.18 & \multicolumn{1}{c|}{45.25} & 24.19 & 24.50 & \multicolumn{1}{c|}{38.49} & 12.74 & 10.92 & \multicolumn{1}{c|}{25.42} & 10.92 & 10.99 & \multicolumn{1}{c|}{8.03} & 24.89 & 23.99 & 36.12 \\
\multicolumn{1}{c|}{FedAWA} & \multicolumn{1}{c|}{CVPR'25} & 46.86 & 41.91 & \multicolumn{1}{c|}{70.31} & 36.82 & 36.65 & \multicolumn{1}{c|}{54.83} & 25.83 & 28.62 & \multicolumn{1}{c|}{43.67} & 17.39 & 21.82 & \multicolumn{1}{c|}{28.76} & 12.35 & 12.18 & \multicolumn{1}{c|}{11.08} & 27.85 & 28.24 & 41.73 \\
\multicolumn{1}{c|}{Fed-RAM} & \multicolumn{1}{c|}{CVPR'25} & 45.53 & 39.83 & \multicolumn{1}{c|}{\underline{72.06}} & \underline{37.81} & \underline{37.01} & \multicolumn{1}{c|}{59.51} & \underline{26.86} & \underline{26.07} & \multicolumn{1}{c|}{\underline{51.93}} & \underline{20.92} & \underline{22.03} & \multicolumn{1}{c|}{27.50} & \underline{11.60} & \underline{13.91} & \multicolumn{1}{c|}{\underline{12.25}} & \underline{28.54} & \underline{27.77} & \underline{44.65} \\
\multicolumn{1}{c|}{FedMVP} & \multicolumn{1}{c|}{ICCV'25} & \underline{48.06} & \underline{42.28} & \multicolumn{1}{c|}{71.44} & 36.06 & 36.38 & \multicolumn{1}{c|}{\underline{59.72}} & 25.61 & 26.00 & \multicolumn{1}{c|}{45.46} & 15.04 & 16.64 & \multicolumn{1}{c|}{\underline{28.54}} & 10.18 & 12.04 & \multicolumn{1}{c|}{10.38} & 26.99 & 26.67 & 43.11 \\
\rowcolor{gray!10}\multicolumn{1}{c|}{FedMPT} & \multicolumn{1}{c|}{Ours} & \textbf{51.72} & \textbf{45.16} & \multicolumn{1}{c|}{\textbf{72.88}} & \textbf{40.15} & \textbf{40.66} & \multicolumn{1}{c|}{\textbf{63.72}} & \textbf{30.42} & \textbf{32.83} & \multicolumn{1}{c|}{\textbf{56.64}} & \textbf{24.47} & \textbf{26.27} & \multicolumn{1}{c|}{\textbf{31.05}} & \textbf{15.81} & \textbf{18.97} & \multicolumn{1}{c|}{\textbf{15.01}} & \textbf{32.51} & \textbf{32.78} & \textbf{47.86} \\
\rowcolor{gray!10}\multicolumn{1}{c|}{\textbf{$\Delta$ Prev. Best}} & \multicolumn{1}{c|}{\textbackslash{}} & \textcolor{red}{\textbf{+3.12}} & \textcolor{red}{\textbf{+2.88}} & \multicolumn{1}{c|}{\textcolor{red}{\textbf{+0.82}}} & \textcolor{red}{\textbf{+2.34}} & \textcolor{red}{\textbf{+3.65}} & \multicolumn{1}{c|}{\textcolor{red}{\textbf{+4.00}}} & \textcolor{red}{\textbf{+3.56}} & \textcolor{red}{\textbf{+4.07}} & \multicolumn{1}{c|}{\textcolor{red}{\textbf{+4.71}}} & \textcolor{red}{\textbf{+3.55}} & \textcolor{red}{\textbf{+4.24}} & \multicolumn{1}{c|}{\textcolor{red}{\textbf{+2.29}}} & \textcolor{red}{\textbf{+3.46}} & \textcolor{red}{\textbf{+5.06}} & \multicolumn{1}{c|}{\textcolor{red}{\textbf{+2.12}}} & \textcolor{red}{\textbf{+3.97}} & \textcolor{red}{\textbf{+4.54}} & \textcolor{red}{\textbf{+3.21}} \\ \hline
\end{tabular}\label{tab:22}
} 
\end{center}
\vspace{-0.8cm}
\end{table*}

\vspace{-0.4cm}
\paragraph{Implementation Details.} We employ CLIP (ViT-B/16) with both encoders frozen and the SGD optimizer with a maximum learning rate of 0.001. The batch-size is 32. $\lambda$ is 0.2. $\tau\text{=}4$. The length of learnable tokens for conditions and classes $\beta_{cond}$, $\beta_{cls}$ is 4. The training epoch for VOC2007 and Multi-scene is 100; For COCO2014, NUS-Wide, and MLRSNet, it's 200. A communication round is conducted after one epoch by default. \textit{Epoch and round settings are the same for all methods for fairness.} The client number $S$ and participation rate $\epsilon$ vary in different experiments.

\begin{table}[t]
\caption{Results on the real-world MLR Benchmark. We report the mAP, CF1, and OF1. The best results are marked with \textbf{bold}.}
\vspace{-0.5cm}
\begin{center} 
\renewcommand{\arraystretch}{1}   
\setlength\tabcolsep{8pt} 
\resizebox{1\linewidth}{!}{
\begin{tabular}{c|ccc|ccc}
\hline
 & \multicolumn{3}{c|}{Multi-Scene} & \multicolumn{3}{c}{MLRSNet} \\ \cline{2-7} 
\multirow{-2}{*}{Method} & mAP & CF1 & OF1 & mAP & CF1 & OF1 \\ \hline
Fed-DualCoOp & 40.09 & 31.37 & 50.18 & 38.24 & 40.61 & 66.97 \\
Fed-MaPLe & 33.18 & 30.02 & 29.70 & 37.37 & 46.37 & 61.18 \\
FedPGP & 45.96 & 35.85 & 53.77 & 52.75 & 45.27 & 50.75 \\
Fed-TCP & 45.25 & 37.43 & 53.43 & 43.01 & 42.32 & 67.61 \\
FedTPG & 44.45 & 35.55 & 52.39 & 31.38 & 35.11 & 62.65 \\
Fed-PosCoOp & 42.30 & 35.06 & 47.79 & 37.25 & 39.89 & 65.14 \\
FedAWA & 47.58 & 37.92 & 53.53 & 40.89 & 42.32 & 68.13 \\
Fed-RAM & 49.41 & 39.07 & 52.70 & \underline{47.83} & \underline{46.07} & 66.18 \\
FedMVP & \underline{49.56} & \underline{39.62} & \underline{54.16} & 45.89 & 44.98 & \underline{66.52} \\
\rowcolor{gray!10}FedMPT & \textbf{53.68} & \textbf{43.97} & \textbf{57.83} & \textbf{58.76} & \textbf{50.86} & \textbf{71.22} \\
\rowcolor{gray!10}\textbf{$\Delta$ Prev. best} & \textcolor{red}{\textbf{+4.12}} & \textcolor{red}{\textbf{+4.35}} & \textcolor{red}{\textbf{+3.67}} & \textcolor{red}{\textbf{+6.01}} & \textcolor{red}{\textbf{+4.49}} & \textcolor{red}{\textbf{+3.09}} \\ \hline
\end{tabular}\label{tab:33} 
} 
\end{center}
\vspace{-0.7cm}
\end{table}

\subsection{Experiment Results}
\noindent\textbf{Results of Heterogeneity Benchmark.} We report the results in Table \ref{tab:11}, where we draw the following key observations: \textbf{(a)} directly transferring standard prompt learning methods like TCP and MaPLe yields unsatisfying performance and robustness to heterogeneity changes (a maximal degradation of about 8.14\% in mAP), potentially stemming from their neglect of contextual region semantics; \textbf{(b)} although applying MLR methods to federated scenario yields competitive performance, they're comparably vulnerable to increased heterogeneity for severe overfitting to local data; \textbf{(c)} SOTAs of federated learning like FedTPG and FedMVP, generally achieve top-tier performance (82.72\%, 82.43\%) under severe heterogeneity, but approaches MLR methods in average metrics for their suboptimal multi-label modeling capabilities. In contrast, our FedMPT indisputably outperforms them by a substantial margin (mAP: 3.84\% on VOC2007, 3.01\% on COCO2014, 3.36\% on NUS-Wide). Meanwhile, FedMPT shows less fluctuation faced with data heterogeneity, highlighting its robustness and superiority.

\noindent\textbf{Results of Federated Part-annotation Benchmark.} The results are shown in Table \ref{tab:22}. We find that existing SOTAs like Fed-RAM and FedMVP, which excel on fully-annotated data, generally yield much degraded performance when the annotation mask increases, validated by their average decrease of about 20\%, 32\%, and 38\% in three datasets, respectively. We argue that since these methods rely solely on coarse-grained object categories for cross-modal matching, they overemphasize the adaptation of individual prompts to local data distributions, thereby impairing the generalization capability of the global model. In contrast, our FedMPT consistently outperforms existing methods with its decomposition of conditions: on average of three datasets, FedMPT surpasses existing best results by about 5.3\%, 5.8\%, and 4\% at three metrics; the benefits from FedMPT show a positive correlation with $\texttt{mask}$ (2.26\%$\rightarrow$$7.25\%$ on COCO2014), verifying its robustness.

\noindent\textbf{Results of Federated Real-world MLR Benchmark.} In Table \ref{tab:33} we report the results on two real-world MLR datasets. While real-world datasets present more noise and tricky instances, the reliance on multiple conditions makes our FedMPT more robust to potential noise in the data compared to existing methods. Concretely, we observe that FedMPT keeps its superiority and outperforms existing SOTAs by 4.27\% mAP on Multi-Scene and 6.01\% mAP in MLRSNet, highlighting its  promising versatility.

\begin{table}[t]
\caption{Ablations on different proposed modules.}
\vspace{-0.6cm}
\begin{center} 
\renewcommand{\arraystretch}{1}   
\setlength\tabcolsep{8pt} 
\resizebox{1\linewidth}{!}{
\begin{tabular}{cccc|cccc}
\hline
CPs & Adapters & OT & Gate & mAP & CF1 & OF1 & Avg. \\ \hline
\Checkmark &  &  &  & 87.08 & 82.90 & 81.66 & 83.88 \\
 & \Checkmark &  &  & 84.40 & 78.31 & 80.04 & 80.92 \\
\Checkmark & \Checkmark &  &  & 87.62 & 83.05 & 83.68 & 84.78 \\
\Checkmark &  & \Checkmark &  & 89.35 & 83.82 & 82.53 & 85.23 \\
\Checkmark & \Checkmark & \Checkmark &  & 89.64 & 83.75 & 83.72 & 85.70 \\
\Checkmark & \Checkmark &  & \Checkmark & 87.89 & 83.34 & 83.01 & 84.75 \\
\rowcolor{gray!10}
\Checkmark & \Checkmark & \Checkmark & \Checkmark & 90.10 & 83.96 & 84.50 & 86.19 \\ \hline
\end{tabular}}\label{tab:66} 
\end{center}
\vspace{-0.7cm}
\end{table}

\begin{table}[t]
\caption{Comparison of computation overhead on VOC2007.}
\vspace{-0.6cm}
\begin{center} 
\renewcommand{\arraystretch}{1}   
\setlength\tabcolsep{5pt} 
\resizebox{1\linewidth}{!}{
\begin{tabular}{c|c|c|c|c}
\hline
Method & \# Total Param. & \# Tunable Param. & Training Time & mAP \\ \hline
Fed-PosCoOp & 86.60 M  & 0.02 M & 38.93 ms/iter & 84.46 \\
Fed-MaPLe & 90.14 M & 3.56 M & 65.25 ms/iter & 81.87 \\
FedTPG & 90.79 M & 4.21 M & 59.43 ms/iter & 84.23 \\
Fed-RAM & 99.60 M & 13.02 M & 384.51 ms/iter & 85.54 \\
FedMVP & 87.72 M & 1.14 M & 75.80 ms/iter & 85.61 \\
\rowcolor{gray!10}FedMPT & 87.38 M & 0.80 M & 97.03 ms/iter & 90.10 \\ \hline
\end{tabular}}\label{tab:55} 
\end{center}
\vspace{-0.8cm}
\end{table}

\section{Ablation Studies and Discussions}
\noindent\textbf{Proposed Modules.} The results are shown in Table \ref{tab:66}. We find that Condition Prompts (CPs) provide a more substantial performance gain than the adapters alone, underscoring the critical role of explicit condition modeling over mere visual feature adaptation. Moreover, employing OT delivers an average enhancement of  1.44\% mAP. However, employing gating without OT has more limited improvement (+0.27\% mAP) than that when OT is applied (+2.21\% mAP), indicating that the synergistic effects between conditions rely on OT to mediate trade-offs among patches. 

\noindent\textbf{Cost Analysis.} Table \ref{tab:55} reports the computation overheads. Fed-PosCoOp yields the least learnable parameters but also the worst performance; FedMVP and FedRAM yield comparable performance (85.61\% and 85.54\% mAP), but at the cost of significantly expanded training time and parameters. Comparably, FedMPT achieves the best performance with minor extra overhead, showing its efficiency. 
 
\begin{figure}[t] 
    \centering
    \includegraphics[width=1\linewidth]{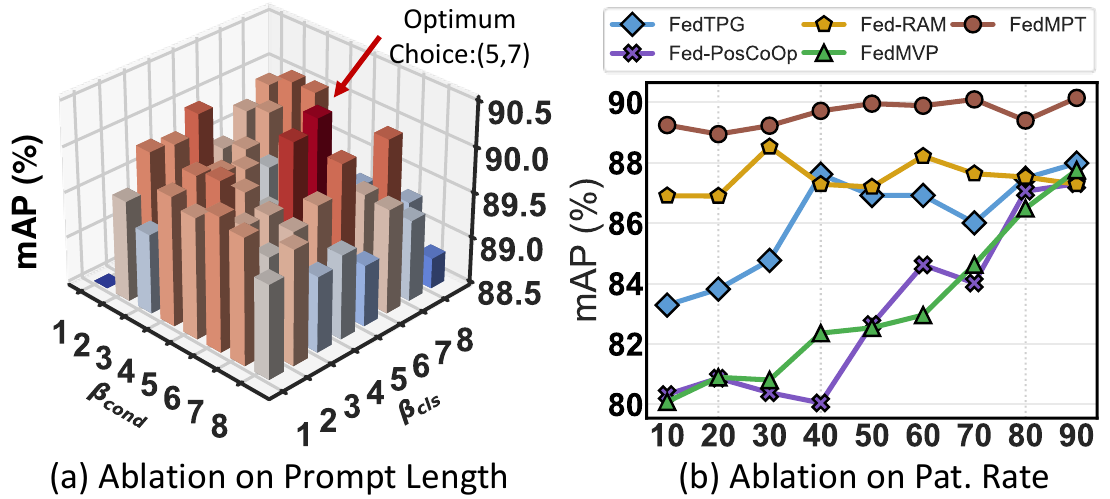} 
    \vspace{-0.6cm}
    \caption{Ablation studies on prompt length and participation rate.} 
    \vspace{-0.7cm}
    \label{fig:66}
\end{figure}

\noindent\textbf{Number of Learnable Tokens $\beta_{cond}$ and $\beta_{cls}$.} $\beta_{cond}$ and $\beta_{cls}$ control the number of learnable tokens of conditions and classes, respectively. We perform an exemplar grid-search to them on VOC2007 in Figure \ref{fig:66}.a. While we can observe that less learnable tokens tend to result in poor performance, excessively large choices also cause minor degradation, where possible reasons lie in the increased learning difficulty of the prompts. The optimal choice is (5,7), which is a trade-off between the capability and difficulty.

\begin{figure}[t] 
    \centering
    \includegraphics[width=1\linewidth]{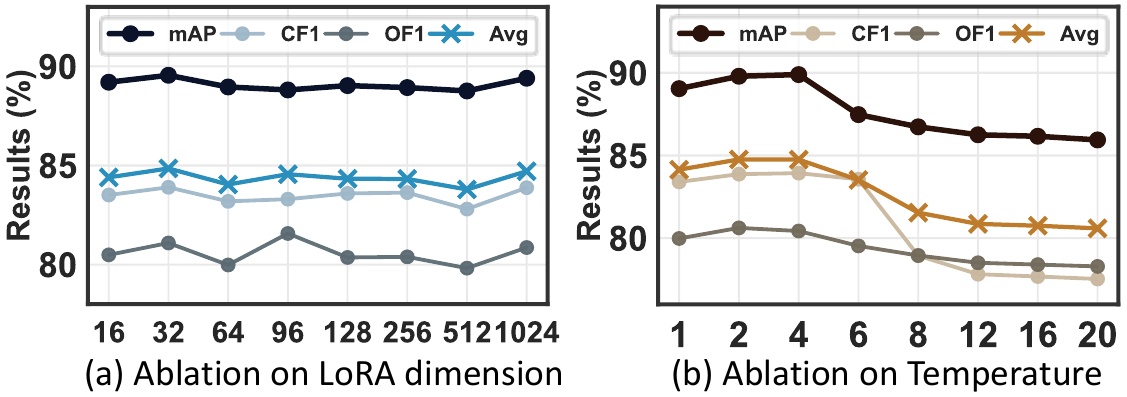} 
    \vspace{-0.8cm}
    \caption{Ablation studies on LoRA dimension and temperature.} 
    \vspace{-0.3cm}
    \label{fig:55}
\end{figure}  

\noindent\textbf{LoRA Dimension $D_s$ and Temperature $\tau$.} We separately alter $D_s$ and $\tau$ to $[16\text{-}1024]$ and $[1\text{-}20]$ and investigate the mAP in Figure \ref{fig:55}. \textbf{For $D_s$}, we find that its change brings a minor fluctuation generally and a continual decrement as $D_s$ grows larger than 32, possibly from overfitting. $D_s=32$ is the optimal choice. \textbf{For $\tau$}, the results show an obvious degradation of all metrics when it grows larger than $4$, probability due to the indistinguishability between classes. Our experiments show that $\tau=4$ is the optimal.

\noindent\textbf{Participation Rate of Clients.}  We vary the participation rate $\epsilon$ from 10\% to 90\% and report the results in Figure \ref{fig:66}.b. We can observe that FedMPT consistently  outperforms other methods and exemplifies a gentle change under different participation rates.  In contrast, some methods like FedMVP and FedTPG suffer from dramatic degradation ($\sim$5\%/7\% mAP) when participation rate decreases.

\section{Conclusion}
The integration of Multi-Label Recognition (MLR) with Federated Learning (FL) introduces a significant risk of overfitting to spurious local label correlations. To address this, we present FedMPT, a novel framework grounded in causal analysis that leverages conditions to approximate true class relationships. FedMPT employs an LLM-driven pipeline to generate abstract conditions, aligns them with visual regions via optimal transport, and  integrates their contributions through a gating mechanism. Experiments validate the superiority across federated benchmarks.  

\section*{Acknowledgement}
The authors gratefully acknowledge the support from the National Natural Science Foundation of China (NSFC) under Grant Nos. 62402472, and 12227901. This work was also supported by the Natural Science Foundation of Jiangsu Province (No. BK20240461), the Project of Stable Support for Youth Team in Basic Research Field, CAS (No. YSBR-005), and the Academic Leaders Cultivation Program at USTC. The AI-driven experiments, simulations and model training were performed on the robotic AI-Scientist platform of Chinese Academy of Sciences.

{
    \small
    \bibliographystyle{ieeenat_fullname}
    \bibliography{main}
}


\clearpage
\newpage

\appendix
\section{More Experiments}
\subsection{More Ablation Studies on Participation Rate}
Table \ref{tab:a11} presents extended ablation studies on all baselines. FedMPT consistently outperforms all state-of-the-art methods by substantial margins, achieving gains of 2.22\% mAP, 2.88\% CF1, and 3.26\% OF1. Notably, methods relying more heavily on visual adaptation (e.g., FedMVP and Fed-MaPLe) exhibit significantly higher performance variance as the client participation rate decreases. This can be attributed to their local models' heightened susceptibility to overfitting client-specific data; when aggregated under low participation rates, the global model is disproportionately influenced by these overfitted local models, making it more vulnerable to heterogeneity and distribution shifts.

\subsection{Experiments on ZSL and GZSL Benchmarks}
Following RAM~\cite{tan2025recover}, we conduct more experiments on two other benchmarks~\cite{sun2022dualcoop,hu2023dualcoop++}: the Federated Zero-shot Generalization Benchmark (FZSL) and the Federated Generalized Zero-shot Generalization Benchmark (FGZSL). These two benchmarks evaluate the model's robustness to unseen classes, which is a comparably harsher setting.

\paragraph{Benchmarks Overview.}
The Federated Zero-shot Generalization Benchmark (ZSL) first splits all classes into seen and unseen classes, then performs clustering on training samples and sends each cluster to a client as its private data. Local models are trained on their private data, with only the seen classes annotated. The global model is evaluated on the test data, with only the unseen classes considered in all of the metrics. The Federated Generalized Zero-shot Generalization Benchmark (FGZSL) is similar, but the global model is evaluated on the test data, with both seen and unseen classes considered in all of the metrics. COCO2014~\cite{lin2014microsoft} and NUS-Wide~\cite{chua2009nus} are employed for the above two benchmarks. For COCO2014, the dataset is split into 48 seen classes and 17 unseen classes. NUS-WIDE is split into 925 seen classes and 81 unseen classes.

\begin{table*}[t]
\caption{\textbf{More ablation studies of FedMPT and other methods on the participation rate} . We report the mAP, CF1 and OF1 with the part-annotation setting \texttt{Mask} varying from 10\% to 90\%. The best results are marked with \textbf{bold}.}
\vspace{-0.5cm}
\begin{center} 
\renewcommand{\arraystretch}{1}   
\setlength\tabcolsep{3pt} 
\resizebox{1\textwidth}{!}{
\begin{tabular}{c|c|ccc|ccc|ccc|ccc|ccc|ccc}
\hline
 &  & \multicolumn{3}{c|}{Pat. rat.=0.1} & \multicolumn{3}{c|}{Pat. rat.=0.3} & \multicolumn{3}{c|}{Pat. rat.=0.5} & \multicolumn{3}{c|}{Pat. rat.=0.7} & \multicolumn{3}{c|}{Pat. rat.=0.9} & \multicolumn{3}{c}{Avg} \\ \cline{3-20} 
\multirow{-2}{*}{Method} & \multirow{-2}{*}{Venues} & mAP & CF1 & OF1 & mAP & CF1 & OF1 & mAP & CF1 & OF1 & mAP & CF1 & OF1 & mAP & CF1 & OF1 & mAP & CF1 & OF1 \\ \hline
Fed-DualCoOp & NeurIPS'22 & 85.92 & \underline{79.88} & 77.74 & 83.59 & 77.57 & 76.36 & 84.70 & 77.85 & 75.73 & 84.27 & 76.95 & 78.13 & 85.44 & 78.27 & 76.98 & 84.78 & 78.10 & 76.99 \\
Fed-SCPNet & CVPR'23 & 79.67 & 68.96 & 72.86 & 76.38 & 69.29 & 69.98 & 77.09 & 71.60 & 70.71 & 78.96 & 70.70 & 72.17 & 80.44 & 77.37 & 74.25 & 78.51 & 71.58 & 71.99 \\
Fed-MaPLe & CVPR'23 & 82.48 & 75.60 & 73.68 & 83.18 & 77.46 & 72.99 & 82.81 & 77.54 & 75.03 & 87.10 & 81.46 & 78.48 & 86.25 & 81.25 & 79.87 & 84.36 & 78.66 & 76.01 \\
FedPGP & ICML'24 & 79.43 & 72.56 & 68.38 & 78.68 & 73.29 & 69.40 & 81.18 & 75.16 & 65.22 & 81.57 & 74.17 & 66.58 & 83.55 & 76.60 & 72.77 & 80.88 & 74.36 & 68.47 \\
Fed-TCP & CVPR'24 & 79.66 & 73.58 & 74.14 & 81.19 & 75.31 & 72.07 & 81.83 & 76.95 & 70.63 & 82.07 & 76.00 & 69.15 & 83.00 & 77.80 & 79.12 & 81.55 & 75.93 & 73.02 \\
FedTPG & ICLR'24 & 83.29 & 75.02 & 76.83 & 84.73 & 76.01 & 78.23 & 86.92 & 80.47 & 79.95 & 86.10 & 80.64 & 77.26 & 87.89 & 80.12 & 77.62 & 85.79 & 78.45 & 77.98 \\
Fed-PosCoOp & WACV'25 & 80.33 & 73.63 & 78.62 & 80.39 & 75.49 & 79.69 & 82.65 & 76.04 & 75.27 & 84.02 & 78.97 & 68.99 & 87.31 & 83.29 & 80.57 & 82.94 & 77.48 & 76.63 \\
FedAWA & CVPR'25 & 79.98 & 70.04 & 78.11 & 78.33 & 68.31 & 69.77 & 80.09 & 71.63 & 70.30 & 84.45 & 72.93 & 70.97 & 83.82 & 75.31 & 72.73 & 81.33 & 71.64 & 72.38 \\
Fed-RAM & CVPR'25 & \underline{86.91} & 79.09 & \underline{80.05} & \underline{88.53} & \underline{82.13} & \underline{82.56} & \underline{87.20} & \underline{81.45} & \underline{80.40} & \underline{87.64} & \underline{80.70} & \underline{81.84} & 87.29 & 80.85 & 78.29 & \underline{87.51} & \underline{80.84} & \underline{80.63} \\
FedMVP & ICCV'25 & 80.09 & 72.73 & 72.80 & 80.81 & 72.50 & 72.12 & 82.36 & 72.23 & 70.88 & 84.63 & 77.93 & 78.03 & \underline{87.76} & \underline{82.88} & \underline{80.00} & 83.13 & 75.65 & 74.77 \\
FedMPT & Ours & \textbf{89.25} & \textbf{82.41} & \textbf{84.78} & \textbf{89.23} & \textbf{82.97} & \textbf{84.05} & \textbf{89.96} & \textbf{83.74} & \textbf{83.25} & \textbf{90.10} & \textbf{84.51} & \textbf{83.30} & \textbf{90.15} & \textbf{84.97} & \textbf{84.05} & \textbf{89.74} & \textbf{83.72} & \textbf{83.89} \\ \hline
\textbf{$\Delta$ Prev. Best} & \textbackslash{} & \textcolor{red}{ \textbf{+2.34}} & \textcolor{red}{ \textbf{+2.53}} & \textcolor{red}{ \textbf{+4.73}} & \textcolor{red}{ \textbf{+0.70}} & \textcolor{red}{ \textbf{+0.84}} & \textcolor{red}{ \textbf{+1.49}} & \textcolor{red}{ \textbf{+2.76}} & \textcolor{red}{ \textbf{+2.29}} & \textcolor{red}{ \textbf{+2.85}} & \textcolor{red}{ \textbf{+2.46}} & \textcolor{red}{ \textbf{+3.05}} & \textcolor{red}{ \textbf{+1.46}} & \textcolor{red}{ \textbf{+2.26}} & \textcolor{red}{ \textbf{+1.68}} & \textcolor{red}{ \textbf{+3.48}} & \textcolor{red}{ \textbf{+2.22}} & \textcolor{red}{ \textbf{+2.88}} & \textcolor{red}{ \textbf{+3.26}} \\ \hline
\end{tabular}\label{tab:a11}
} 
\end{center}
\vspace{-0.8cm}
\end{table*}

\begin{table}[t]
\caption{\textbf{Results on the FZSL benchmark.} We report the mAP, CF1 and OF1. The best results are marked with \textbf{bold}.}
\vspace{-0.5cm}
\begin{center} 
\renewcommand{\arraystretch}{1}   
\setlength\tabcolsep{6pt} 
\resizebox{1\linewidth}{!}{
\begin{tabular}{c|ccc|ccc}
\hline
 & \multicolumn{3}{c|}{COCO2014} & \multicolumn{3}{c}{NUS-WIDE} \\ \cline{2-7} 
\multirow{-2}{*}{Method} & mAP & CF1 & OF1 & mAP & CF1 & OF1 \\ \hline
Fed-DualCoOp & 11.19 & 8.59 & \underline{10.97} & 48.65 & 50.39 & 67.52 \\
Fed-SCPNet & 5.89 & 5.34 & 5.71 & 38.35 & 41.62 & 61.61 \\
Fed-MaPLe & 2.63 & 2.95 & 2.28 & 53.13 & 53.28 & 68.65 \\
FedPGP & 6.13 & 4.98 & 4.90 & 51.10 & 53.07 & 67.72 \\
Fed-TCP & 9.40 & 8.97 & 9.99 & 42.90 & 44.77 & 64.30 \\
FedTPG & 10.53 & 9.90 & 8.37 & 42.14 & 40.36 & 65.30 \\
Fed-PosCoOp & 10.68 & 8.22 & 8.17 & 47.44 & 46.03 & 64.47 \\
FedAWA & \underline{14.88} & \underline{10.56} & 8.64 & 44.40 & 47.78 & 64.75 \\
Fed-RAM & 13.84 & 8.77 & 9.22 & 50.52 & 42.69 & 66.79 \\
FedMVP & 6.53 & 5.73 & 2.09 & \underline{52.73} & \underline{52.61} & \underline{70.67} \\
FedMPT & \textbf{19.76} & \textbf{16.02} & \textbf{15.38} & \textbf{57.17} & \textbf{56.94} & \textbf{71.83} \\ \hline
\textbf{$\Delta$ Prev. Best} & \textcolor{red}{ \textbf{+4.88}} & \textcolor{red}{ \textbf{+5.46}} & \textcolor{red}{ \textbf{+4.41}} & \textcolor{red}{ \textbf{+4.04}} & \textcolor{red}{ \textbf{+3.66}} & \textcolor{red}{ \textbf{+1.16}} \\ \hline
\end{tabular}\label{tab:a22} 
} 
\end{center}
\vspace{-0.7cm}
\end{table}

\paragraph{Results on FZSL Benchmark.} As shown in Table \ref{tab:a22}, all methods exhibit significant performance degradation on the challenging COCO2014 benchmark. For instance, Fed-MaPLe achieves only 2.63 mAP, while FedMVP reaches 6.53 mAP. These results underscore the particular difficulty of achieving robust class-level generalization in federated learning environments. In contrast, FedMPT substantially outperforms all SOTA methods across both datasets, demonstrating superior generalization and robustness.

\paragraph{Results on FGZSL Benchmark.} As shown in Table \ref{tab:a33}, the performance gain of FedMPT is consistent, as it outperforms existing SOTAs by 3.79 mAP and 3.91 mAP on COCO2014 and NUS-Wide, respectively. This result demonstrates FedMPT's substantial generalization capabilities across both seen and unseen categories.

\subsection{Experiments of convergence speed and statistical significance}
The experiment results of convergence speed and statistical significance are shown in Table \ref{tab:sd} and Table \ref{tab:sign}.

\begin{table}[h]
\caption{The convergence speed (x-axis: communication round).}
\vspace{-0.7cm}
\begin{center}
\renewcommand{\arraystretch}{1}   
\setlength\tabcolsep{3pt} 
\resizebox{1\linewidth}{!}{
\begin{tabular}{c|cccccccccccc}
\hline
Method $\vert$ VOC & 1 & 5 & 10 & 20 & 30 & 40 & 50 & 60 & 70 & \multicolumn{1}{l}{80} & \multicolumn{1}{l}{90} & \multicolumn{1}{l}{100} \\ \hline
Fed-DualCoOp & 59.72 & 76.28 & 80.01 & 81.66 & 82.65 & 83.31 & 83.74 & 84.05 & 84.26 & 84.38 & 84.40 & 84.41 \\
FedRAM & 61.64 & 77.50 & 81.11 & 82.29 & 83.65 & 84.52 & 85.04 & 85.31 & 85.55 & 85.67 & 85.68 & 85.68 \\
FedMVP & 53.49 & 76.39 & 78.54 & 80.13 & 81.90 & 82.36 & 83.09 & 83.92 & 84.64 & 85.18 & 85.52 & 85.61 \\
FedMPT & 67.39 & 80.11 & 85.76 & 87.32 & 88.63 & 89.20 & 89.71 & 89.98 & 90.00 & 90.03 & 90.04 & 90.04 \\ \hline
\end{tabular}}\label{tab:sd} 
\end{center}
\vspace{-0.5cm}
\end{table}

\begin{table}[h]
\caption{The effects of gating (upper) and significance test (lower).}
\vspace{-0.7cm}
\begin{center}
\renewcommand{\arraystretch}{1}   
\setlength\tabcolsep{3pt} 
\resizebox{1\linewidth}{!}{
\begin{tabular}{c|ccc|ccc}
\hline
\multirow{2}{*}{Metric} & \multicolumn{3}{c|}{VOC2007} & \multicolumn{3}{c}{COCO2014} \\ \cline{2-7} 
 & mAP & CF1 & OF1 & mAP & CF1 & OF1 \\ \hline
no-gating's SD. & $\pm$1.12 & $\pm$0.73 & $\pm$1.31 & $\pm$0.38 & $\pm$0.40 & $\pm$0.53 \\ \hline
FedMPT's SD. & $\pm$0.27 & $\pm$0.48 & $\pm$0.50 & $\pm$0.22 & $\pm$0.29 & $\pm$0.39 \\ \hline
\hline
$p$-value & $1.907e^{-6}$ & $1.907e^{-6}$ & $1.907e^{-6}$ & $1.907e^{-6}$ & $1.907e^{-6}$ & $1.907e^{-6}$ \\ \hline
Cliff's Delta & 1.000 & 1.000 & 1.000 & 1.000 & 0.990 & 1.000 \\ \hline
\end{tabular}}\label{tab:sign} 
\end{center}
\vspace{-0.8cm}
\end{table}

\section{Ablation Studies on Hyper-parameters}
\begin{figure}[t] 
    \centering
    \includegraphics[width=1\linewidth]{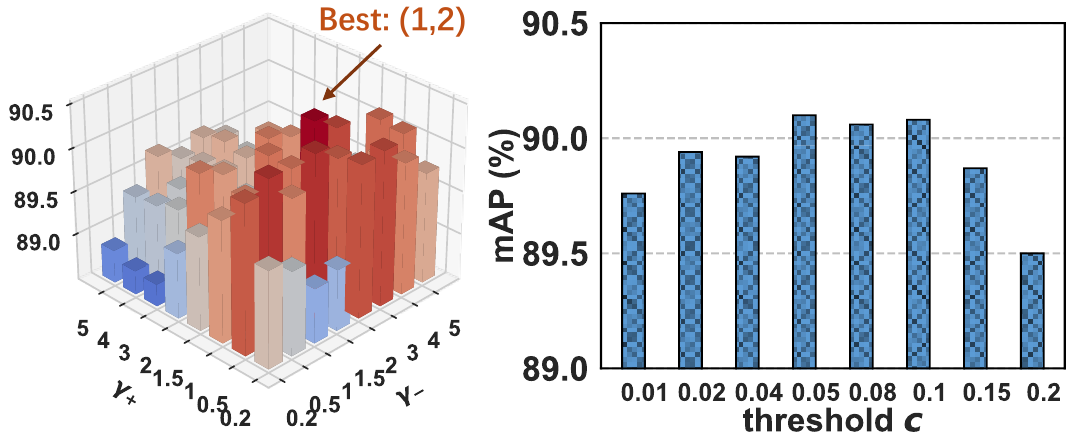} 
    \vspace{-0.7cm}
    \caption{Ablation on \textbf{(left):} $\gamma_{-}$ / $\gamma_{+}$ and \textbf{(right):} $c$.} 
    \label{fig:os} 
\end{figure}

\paragraph{$\gamma+$ and $\gamma-$.} We report the experiment results in Figure \ref{fig:os} (left). We find that both excessively small and large values lead to performance degradation, potentially due to under-constrained/over-constrained optimization for easy-negative samples. Based on experimental results, we selected the values (1,2) for ($\gamma+$, $\gamma-$).

\paragraph{The threshold $c$.} This coefficient controls the clip threshold, where logits below it are clamped to 0. We alter $c$ from 0.01 to 0.2 and report the results in \ref{fig:os} (right). We observed that both excessively small and large values lead to performance degradation, likely due to excessive influence from low-confidence tail classes and over-clipped negative logits, respectively. Based on experimental results, we set this parameter to 0.05.

\section{Introduction of datasets and baselines}
\paragraph{Datasets:} We employ VOC2007~\cite{everingham2015pascal}, COCO2014~\cite{lin2014microsoft}, NUS-Wide~\cite{chua2009nus}, Multi-Scene~\cite{hua2021multiscene} and MLRSNet~\cite{qi2020mlrsnet} in our experiments. Details are in the following:
\begin{itemize}
    \item VOC2007 is a commonly-employed dataset in classification and object detection. It contains 9,963 real-world images annotated with 24,640 object instances across 20 different categories, including people, animals, vehicles, and household items.  The dataset supports multi-label classification, detection, segmentation, and even person layout identification (predicting parts of a person). The diversity of scenes make VOC2007 a standard benchmark for evaluating object recognition algorithms.
    \item COCO2014 is a large-scale dataset for object detection, segmentation, and captioning.  It includes over 330,000 images, more than 200,000 of which are labeled, encompassing around 1.5 million object instances. COCO2014 version covers 80 object categories (from a larger set of 91 classes) with per-instance segmentation masks, making it especially useful for precise localization.  Beyond detection and segmentation, COCO2014 also supports captioning (5 captions per image) and keypoint detection (e.g., human pose), facilitating research in richer scene understanding.   COCO is considered one of the most challenging and representative vision benchmarks.  
    \item NUS-WIDE is a large-scale multi-label image dataset derived from Flickr. It comprises 269,648 images annotated with 81 ground-truth “concepts” (e.g., sky, building, person)  plus up to 5,018 user-provided noisy tags.  Since the original Flickr tags are noisy and incomplete, NUS-WIDE poses realistic challenges for multi-label learning, annotation, and retrieval.  In addition, it provides low-level visual features for each image.
\end{itemize}

\begin{table}[t]
\caption{\textbf{Results on the FGZSL benchmark.} We report the mAP, CF1 and OF1. The best results are marked with \textbf{bold}.}
\vspace{-0.5cm}
\begin{center} 
\renewcommand{\arraystretch}{1}   
\setlength\tabcolsep{6pt} 
\resizebox{1\linewidth}{!}{
\begin{tabular}{c|ccc|ccc}
\hline
 & \multicolumn{3}{c|}{COCO2014} & \multicolumn{3}{c}{NUS-Wide} \\ \cline{2-7} 
\multirow{-2}{*}{Method} & MAP & CF1 & OF1 & MAP & CF1 & OF1 \\ \hline
Fed-DualCoOp & 52.34 & 44.34 & 64.27 & 50.31 & 52.86 & 66.86 \\
Fed-SCPNet & 36.07 & 33.31 & 53.00 & 36.50 & 40.76 & 59.23 \\
Fed-MaPLe & 52.73 & 42,56 & 64.24 & 52.09 & 53.50 & 69.00 \\
FedPGP & 43.76 & 37.54 & 58.88 & 44.46 & 47.59 & 63.76 \\
Fed-TCP & 33.94 & 31.33 & 52.42 & 30.05 & 34.36 & 58.21 \\
FedTPG & 41.22 & 38.49 & 53.75 & 32.51 & 36.23 & 56.41 \\
Fed-PosCoOp & 39.19 & 35.92 & 53.96 & 46.57 & 45.43 & 51.78 \\
FedAWA & 36.91 & 32.81 & 53.60 & 41.04 & 44.94 & 61.00 \\
Fed-RAM & 50.72 & 24.25 & 35.00 & 50.72 & 24.25 & 35.00 \\
FedMVP & 43.95 & 36.72 & 60.03 & 48.79 & 51.31 & 67.37 \\
FedMPT & \textbf{56.52} & \textbf{47.92} & \textbf{68.09} & \textbf{56.00} & \textbf{55.83} & \textbf{72.87} \\ \hline
\textbf{$\Delta$ Prev. Best} & \textcolor{red}{\textbf{+3.79}} & \textcolor{red}{\textbf{+3.58}} & \textcolor{red}{\textbf{+3.82}} & \textcolor{red}{\textbf{+3.91}} & \textcolor{red}{\textbf{+2.33}} & \textcolor{red}{\textbf{+3.87}} \\ \hline
\end{tabular}\label{tab:a33} 
} 
\end{center}
\vspace{-0.7cm}
\end{table}

\paragraph{Baselines.} We employed ten baselines: DualCoOp~\cite{sun2022dualcoop}, SCPNet~\cite{ding2023exploring}, PosCoOp~\cite{rawlekar2025positivecoop}, RAM~\cite{tan2025recover}, MaPLe~\cite{khattak2023maple}, TCP~\cite{yao2024tcp}, FedPGP~\cite{cui2024harmonizing}, FedTPG~\cite{qiu2024federated}, FedAWA~\cite{shi2025fedawa} and FedMVP~\cite{singha2025fedmvp}. Details are as follows:
\begin{itemize}
\item DualCoOp~\cite{sun2022dualcoop} is the first approach that leverages pretrained VLMs (specifically, CLIP) for multi-label recognition. It introduces two prompts, named Positive Prompt and Negative Prompt, to reflect the existence and non-existence of a label.
\item SCPNet~\cite{ding2023exploring} (Semantic Correspondence Prompt Network) proposes to extract the structured semantic prior between labels from CLIP via a structured prior prompter. It then fully explores this prior using a cross-modality prompter and a semantic association module to improve the performance of multi-label recognition with incomplete labels.
\item PosCoOp~\cite{rawlekar2025positivecoop} finds that the negative prompt in DualCoOp~\cite{sun2022dualcoop} does not necessarily need to be conditioned on class names. It leaves the negative prompts unconditionally learnable and only generates positive prompts from class names.
\item RAM~\cite{tan2025recover} recovers the local semantics of CLIP in a memory-efficient manner through the Ladder Local Adapter (LLA), which addresses the loss of local information caused by CLIP's global pretraining objectives. It also designs Knowledge-Constrained Optimal Transport (KCOT), formulating region-to-label matching as an optimal transport problem and integrating Label Presence Detection (LPD) and Teacher Knowledge Transfer (TKT) to suppress meaningless matching, thereby improving the performance of open-vocabulary multi-label recognition.
\item MaPLe~\cite{khattak2023maple} introduces multi-modal prompts to both encoders. The text prompts are dynamically generated from visual prompts via a cross-modality projector.
\item TCP~\cite{yao2024tcp} maps class-level textual knowledge into class-aware prompt tokens through the Textual Knowledge Embedding (TKE) module and then injects them into the text encoder. The algorithm optimizes the model using contrastive loss and knowledge-guided consistency loss. Notably, TKE is a plug-and-play design that can be combined with existing prompt tuning methods, achieving high performance while reducing training time.
\item FedTPG~\cite{qiu2024federated} jointly learns a unified prompt generation network across multiple clients under the federated learning framework. This network generates context-aware prompt vectors conditioned on task-related text inputs, enabling efficient generalization to both seen and unseen classes and datasets.
\item FedAWA~\cite{shi2025fedawa} obtains client vectors by calculating the difference between client model and global model parameters. It then adaptively optimizes aggregation weights based on the alignment between these client vectors and the aggregated global vector and introduces a regularization term to ensure training stability, mitigating the problem of data heterogeneity without requiring proxy data.
\item FedMVP~\cite{singha2025fedmvp} proposes to fuse the visual features of images and the textual attribute features of classes via the PromptFormer module to generate multimodal visual prompts, injects them into the vision encoder of CLIP, and trains the model by combining CLIP similarity loss and consistency loss, thereby improving the generalization ability to unseen classes and domains under the federated learning framework.
\end{itemize}

\section{Condition Study}\label{sec4}
\paragraph{Conditions we used in the experiments.}
The conditions we used for 5 datasets are listed in Table \ref{appx:tab:11}.

\begin{table}[t]
\caption{Lists of conditions used for different datasets. }
\vspace{-0.7cm}
\begin{center} 
\renewcommand{\arraystretch}{1}   
\setlength\tabcolsep{5pt} 
\resizebox{0.48\textwidth}{!}{
\begin{tabular}{c|c}
\hline
Dataset & LLM-generated conditions \\ \hline
VOC2007 & {[}``background'', ``position'', ``shape'', ``action''{]} \\
COCO2014 & {[}``background'', ``position'', ``shape'', ``action''{]} \\
NUS-WIDE & {[}``color'', ``texture'', ``shape'', ``action''{]} \\
Multi-Scene & {[}``color'', ``geometry'', ``shape'', ``contrast''{]} \\
MLRSNet & {[}``size'', ``color'', ``shape'', ``texture''{]} \\ \hline
\end{tabular}
\label{appx:tab:11}
} 
\end{center}
\vspace{-0.6cm}
\end{table}

\paragraph{Ablation study of condition number and condition combinations in LLM queries.} While this factor largely depends on the LLM itself and shows considerable uncertainty, we vary it in the following manner and report the results on VOC2007. We keep the first instruction in our Chain-of-Thought mechanism unchanged, i.e., \textcolor{gray}{Please give a detailed description for each possible combination of the following categories in one sentence. Categories: Aeroplane, Bicycle, Bird, Boat, Bottle,...}, then change the required number $K$ (from 1 to 20) in the other instruction: \textcolor{gray}{Given these descriptions, Please summarize {K} distinct and general conditions under which true class correlations can be reliably represented.}. We conduct five independent API calls and report the most frequent conditions generated by the LLM in Table \ref{appx:tab:22}. We observe that when instructing an LLM to generate very few conditions, the resulting conditions tend to be overly broad (e.g., "context"); conversely, requesting an excessive number of conditions yields outputs that are either overly specific or difficult to observe, such as ``perspective'' and "reflectance". Furthermore, we organize all obtained conditions into prompts for training and record their corresponding accuracy scores, also in Table \ref{appx:tab:22}. We find that using either a few broad conditions (like context,``layout'') or overwhelmingly specific conditions yields suboptimal performance. These results suggest that selecting a moderate quantity represents a reasonable trade-off. As can be observed from the table, we set the required condition number in LLM queries to 4 as a trade-off between efficiency and effectiveness.

We also conduct an experiment to discover the inherent efficiency contrasts of different conditions. Based on the conditions we used in our experiments for VOC2007, i.e., \textit{background, position, shape, action}, we ask GPT-4o to generate some similar but comparably vague and non-representative conditions with: \textcolor{gray}{These are some conditions under which true class correlations in multi-label datasets can be reliably represented: \textit{background, position, shape, action}. Now think conversely. For each given condition, please give another condition that is similar in meaning, but under which the true class correlations in multi-label datasets cannot be reliably represented.} Under five independent API queries, GPT-4o mostly generates \textit{pattern, anchor, surface, habits}, which show more incongruity and are harder to perceive. We then gradually add both kinds of conditions to prompts to discover each condition's effects. The results are in Table \ref{appx:tab:33}. We find that our employed generic conditions \textit{background, position, shape, action} consistently yield more performance gains (0.51\%$\sim$1.38\%) than \textit{pattern, anchor, surface, habits}. This experiment primarily verifies that semantic and generalization disparities also exist among LLM-generated conditions. Due to space constraints and research focus considerations, we refrain from further exploring subsequent cleaning of these conditions and simply rely exclusively on the optimal condition identified in our ablation studies for all experiments.

\begin{table}[t]
\caption{Ablation study on different conditions. }
\vspace{-0.7cm}
\begin{center} 
\renewcommand{\arraystretch}{1}   
\setlength\tabcolsep{16pt} 
\resizebox{0.48\textwidth}{!}{
\begin{tabular}{ll|ll}
\hline
\multicolumn{1}{c}{Context} & \multicolumn{1}{c|}{mAP} & \multicolumn{1}{c}{Context} & \multicolumn{1}{c}{mAP} \\ \hline
\textcolor{gray}{null} & 86.48 & \textcolor{gray}{null}  & 86.48 \\
+ background & 88.61 & + pattern & 87.23 \\
+ position & 88.98 & + anchor & 87.78 \\
+ shape & 89.32 & + action & 88.72 \\
+ action & 90.10 & + habits & 88.59 \\ \hline
\end{tabular}
\label{appx:tab:33}
} 
\end{center}
\vspace{-0.6cm}
\end{table}

\begin{table}[t]
\caption{Ablation study on condition orders. }
\vspace{-0.7cm}
\begin{center} 
\renewcommand{\arraystretch}{1}   
\setlength\tabcolsep{16pt} 
\resizebox{0.42\textwidth}{!}{
\begin{tabular}{c|c}
\hline
Condition Order & mAP \\ \hline
{[}background, position, shape, action{]} & 90.10 \\
{[}position, shape, action, background{]} & 89.97 \\
{[}shape, action, background, position{]} & 90.02 \\
{[}action, background, position, shape{]} & 90.08 \\
{[}background, shape, position, action{]} & 90.05 \\
{[}shape, position, action, background{]} & 90.02 \\
{[}position, action, background, shape{]} & 89.89 \\
{[}action, background, shape, position{]} & 90.01 \\ \hline
\end{tabular}
\label{tab:44}
} 
\end{center}
\vspace{-0.8cm}
\end{table}

\paragraph{Ablation study of condition order.} To investigate this factor, we take \textit{background, position, shape, action} and reorder them in the prompts. The results are shown in Table \ref{tab:44}. We can see that changing the order of conditions does not substantially affect the model's performance, but placing \textit{position} at the beginning seems to cause a minor degradation. We suggest that this may result from CLIP focusing more on earlier text tokens than later ones (an inherent bias of CLIP proposed by~\cite{zhang2024long}), and \textit{position} being comparatively harder to perceive than others.


\begin{table*}[t]
\caption{Lists of conditions under varied requirement number in LLM-queries. }
\vspace{-0.7cm}
\begin{center} 
\renewcommand{\arraystretch}{1}   
\setlength\tabcolsep{3pt} 
\resizebox{1\textwidth}{!}{
\begin{tabular}{c|c|c}
\hline
\multicolumn{1}{c|}{Number } & LLM-generated conditions & mAP (\%)\\ \hline
1 & {[}``context''{]} & 87.14  \\ \hline
2 & {[}``context'', ``layout''{]} & 88.77 \\ \hline
3 & {[}``background'', ``position'', ``action''{]} & 89.61 \\ \hline
4 & {[}``background'', ``position'', ``shape'', ``action''{]} & 90.10 \\ \hline
5 & {[}``color'', ``texture'', ``shape'', ``contrast'', ``action''{]} & 90.06\\ \hline
6 & {[}``background'', ``layout'', ``context'', ``proximity'', ``activity'', ``setting''{]}  & 89.98\\ \hline
8 & {[}``Background'', ``Lightness'', ``Color'', ``Texture'', ``Shape'', ``Size'', ``Position'', ``Action''{]} & 90.12 \\ \hline
10 & {[}``Background'', ``Lightness'', ``Texture'', ``Shape'', ``Color'', ``Size'', ``Action'', ``Position'', ``Occlusion'', ``Perspective''{]} & 90.07\\ \hline
\multirow{2}{*}{12} & {[}``background'', ``lightness'', ``color'', ``texture'', ``shape'', ``size'', ``Action'', ``perspective'', ``contrast'', ``position'', \\
 & ``shadow'', ``context''{]}  & 89.13\\ \hline
\multirow{2}{*}{14} & {[}``background'', ``lightness'', ``color'', ``texture'', ``shape'', ``size'', ``Action'', ``pattern'', ``contrast'', ``position'', ``depth'', \\
 & ``orientation'', ``symmetry'', ``context''{]}  & 89.51\\ \hline
\multirow{2}{*}{16} & {[}``background'', ``lightness'', ``contrast'', ``texture'', ``shape'', ``size'', ``color'', ``pattern'', ``orientation'', ``position'', \\
 & ``Action'', ``depth'', ``symmetry'', ``sharpness'', ``transparency'', ``reflectance''{]} & 89.48 \\ \hline
\multirow{2}{*}{18} & {[}``color'', ``texture'', ``shape'', ``size'', ``pattern'', ``contrast'', ``symmetry'', ``depth'', ``perspective'', ``Action'', \\
 & ``transparency'', ``reflection'', ``shadow'', ``focus'', ``alignment'', ``density'', ``complexity'', ``clarity''{]} & 89.22 \\ \hline
\multirow{2}{*}{20} & {[}``background'', ``lightness'', ``color'', ``texture'', ``shape'', ``size'', ``Action'', ``direction'', ``position'', ``depth'', ``perspective'', \\ 
 & ``contrast'', ``pattern'', ``symmetry'', ``occlusion'', ``context'', ``material'', ``reflection'', ``transparency'', ``shadow''{]} & 89.24  \\ \hline
\end{tabular}
\label{appx:tab:22}
} 
\end{center}
\vspace{-0.8cm}
\end{table*}

\section{Limitations and Broader Impacts}
Although employing conditions to intervene in MLR and learn non-spurious correlations is inspiring, not all conditions can be equally perceived by the VLM. Our ablation study in Sec. \ref{sec4} verifies this: some salient visual conditions like pose, color, and size contribute prominently to the overall performance, while others like symmetry or habits are relatively hard to perceive. This also explains why our model's performance degrades when we ask the LLM to summarize an excessive number of conditions and utilize them in the prompts, as a significant portion of these conditions are redundant and ambiguous. Second, simply leveraging a few learnable tokens to learn condition content may be insufficient in modeling capacity (however, expanding the learnable modules may also dramatically increase complexity). We hope future endeavors will focus on generating more robust and less biased conditions, achieving a better trade-off between efficiency and performance.

From another perspective, this paper treats Multi-Label Recognition (MLR) only as a classification task; other MLR tasks like multi-label object detection and semantic segmentation remain unexplored. Whether these tasks would encounter similar performance degradation when combined with FL should be carefully considered.

\end{document}